\documentclass[sigconf]{acmart}
\usepackage{booktabs}
\usepackage{multirow}

\usepackage[ruled,linesnumbered]{algorithm2e}
\usepackage{algpseudocode}
\usepackage{subfigure}
\usepackage{xcolor}
\usepackage[toc,title]{appendix}
\usepackage{enumitem}
\newcommand{\modelname}{CurvGAN}

\AtBeginDocument{%
  \providecommand\BibTeX{{%
    \normalfont B\kern-0.5em{\scshape i\kern-0.25em b}\kern-0.8em\TeX}}}

\copyrightyear{2022} 
\acmYear{2022} 
\setcopyright{acmcopyright}\acmConference[WWW '22]{Proceedings of the ACM Web Conference 2022}{April 25--29, 2022}{Virtual Event, Lyon, France}
\acmBooktitle{Proceedings of the ACM Web Conference 2022 (WWW '22), April 25--29, 2022, Virtual Event, Lyon, France}
\acmPrice{15.00}
\acmDOI{10.1145/3485447.3512199}
\acmISBN{978-1-4503-9096-5/22/04}

\begin{document}

\title{Curvature Graph Generative Adversarial Networks}

\author{Jianxin Li{$^{1,2}$}, Xingcheng Fu{$^{1,2}$}, Qingyun Sun{$^{1,2}$}, Cheng Ji{$^{1,2}$}, Jiajun Tan{$^{1}$}, Jia Wu{$^{3}$}, Hao Peng{$^{1}$}}
\affiliation{
  \institution{
   $^1$
   Beijing Advanced Innovation Center for Big Data and Brain Computing, Beihang University, Beijing 100191, China\\
   $^2$
   School of Computer Science and Engineering, Beihang University, Beijing 100191, China\\
   $^3$
   School of Computing, Macquarie University, Sydney, Australia\\
   }
  \country{}
}

\email{{lijx,fuxc,sunqy,jicheng,penghao}@act.buaa.edu.cn, chiachiun_than@buaa.edu.cn, jia.wu@mq.edu.au}

\renewcommand{\shortauthors}{Jianxin Li and Xingcheng Fu, et al.}

\begin{abstract}
Generative adversarial network (GAN) is widely used for generalized and robust learning on graph data. 
However, for non-Euclidean graph data, the existing GAN-based graph representation methods generate negative samples by random walk or traverse in discrete space, leading to the information loss of topological properties (e.g. hierarchy and circularity). 
Moreover, due to the topological heterogeneity (i.e., different densities across the graph structure) of graph data, they suffer from serious topological distortion problems. 
In this paper, we proposed a novel Curvature Graph Generative Adversarial Networks method, named \textbf{\modelname}, which is the first GAN-based graph representation method in the Riemannian geometric manifold. 
To better preserve the topological properties, we approximate the discrete structure as a continuous Riemannian geometric manifold and generate negative samples efficiently from the wrapped normal distribution. 
To deal with the topological heterogeneity, we leverage the Ricci curvature for local structures with different topological properties, obtaining to low-distortion representations. 
Extensive experiments show that \modelname~consistently and significantly outperforms the state-of-the-art methods across multiple tasks and shows superior robustness and generalization. 
\end{abstract}



\maketitle

\section{Introduction}
Complex networks are widely used to model the complex relationships between objects~\cite{girvan2002community,6729567}, such as social networks~\cite{hamilton2017inductive}, academic networks~\cite{maulik2006advanced,sun2020pairwise}, and biological networks~\cite{gilmer2017neural,sun2021sugar}. 
In recent years, graph representation learning has shown its power in capturing the irregular but related complex structures in graph data~\cite{fanzhenIJCAI20,Su_2021,9565320}. 
The core assumption of graph representation learning is that topological properties are critical to representational capability~\cite{peng2021reinforced,peng2022reinforced}.
As the important topology characteristics, cycle and tree structures are ubiquitous in many real-world graphs, such as the cycle structure of family or friend relationships, the hypernym structure in natural languages~\cite{NickelK17Poincare,PoincareGlove}, the subordinate structure of entities in the knowledge graph~\cite{wang2020h2kgat}, and the cascade structure of information propagation in social networks~\cite{zubiaga2018detection}. 
Since there are many unknowable noises in the real networks, these noises have huge impact on the topology of the network.
Ignoring the existence of noise in the environment leads to the over-fitting problem in the learning process. 
In this paper, we focus on how to learn the generalization and robust representation for a graph.


\begin{figure}[!t]
\centering
\subfigure[Karate Club network.]{
\includegraphics[width=0.30\linewidth]{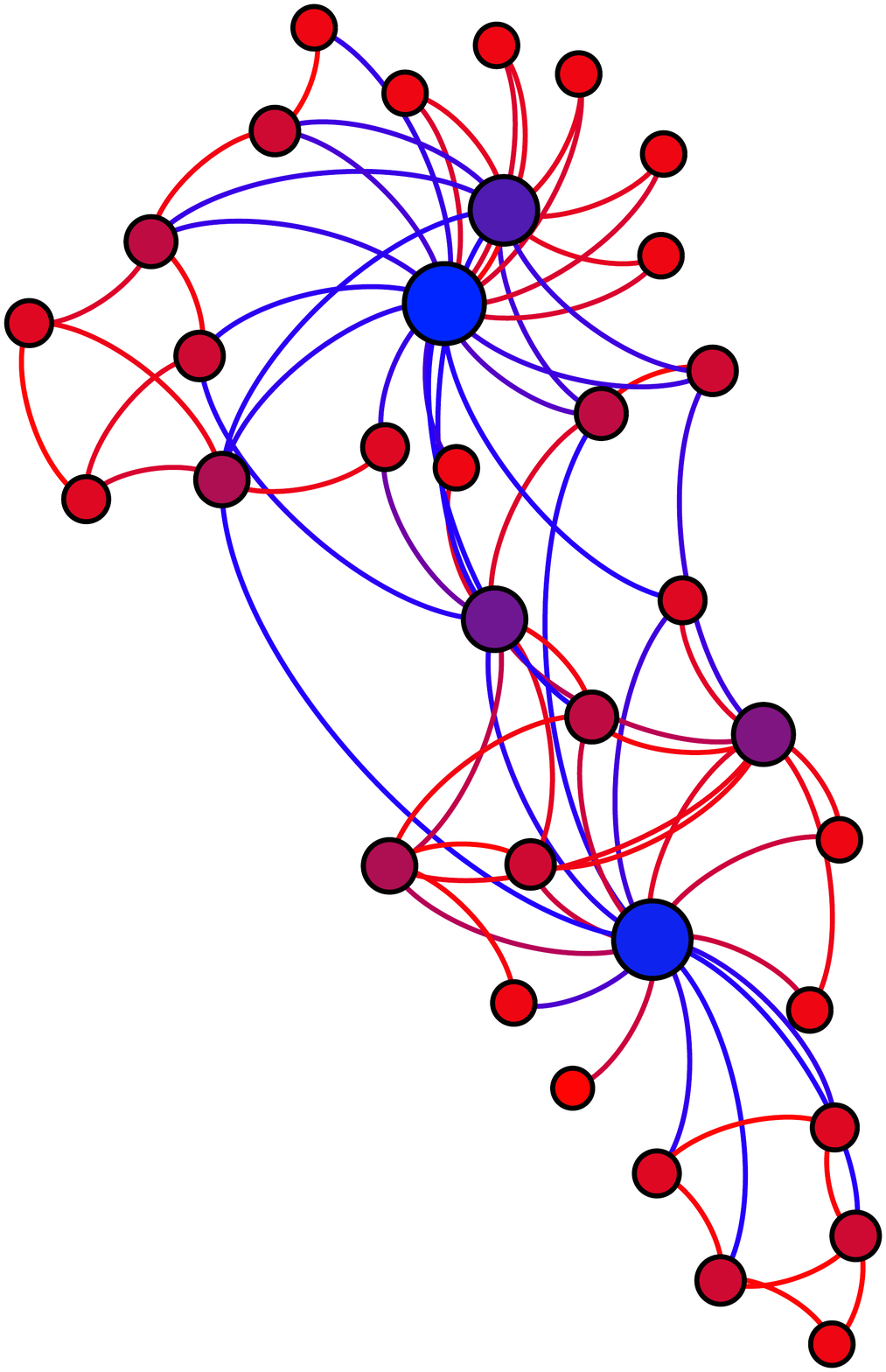}
}%
\subfigure[Tree-like structures.]{
\includegraphics[width=0.30\linewidth]{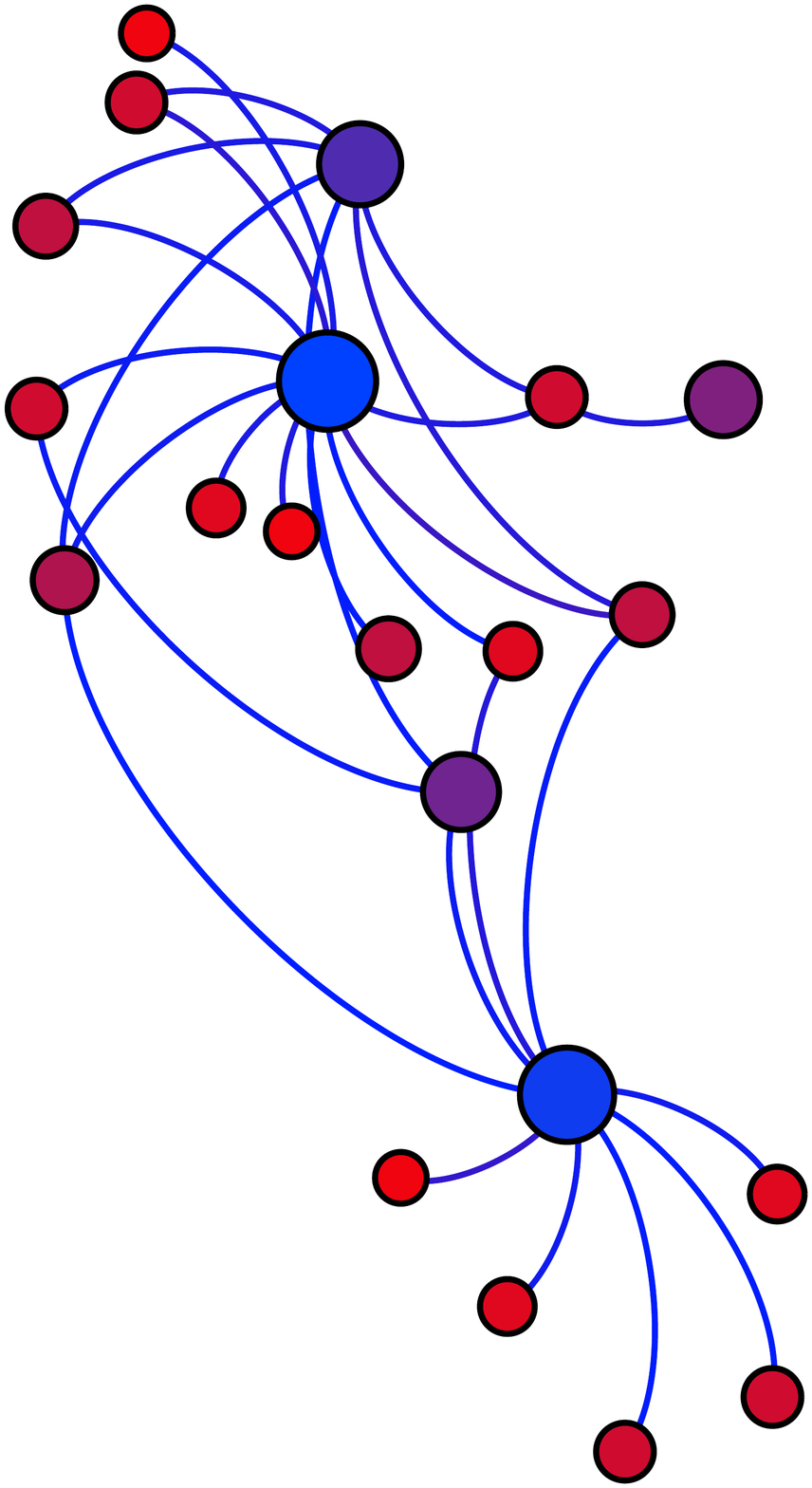}
}%
\subfigure[Cyclic structures.]{
\includegraphics[width=0.30\linewidth]{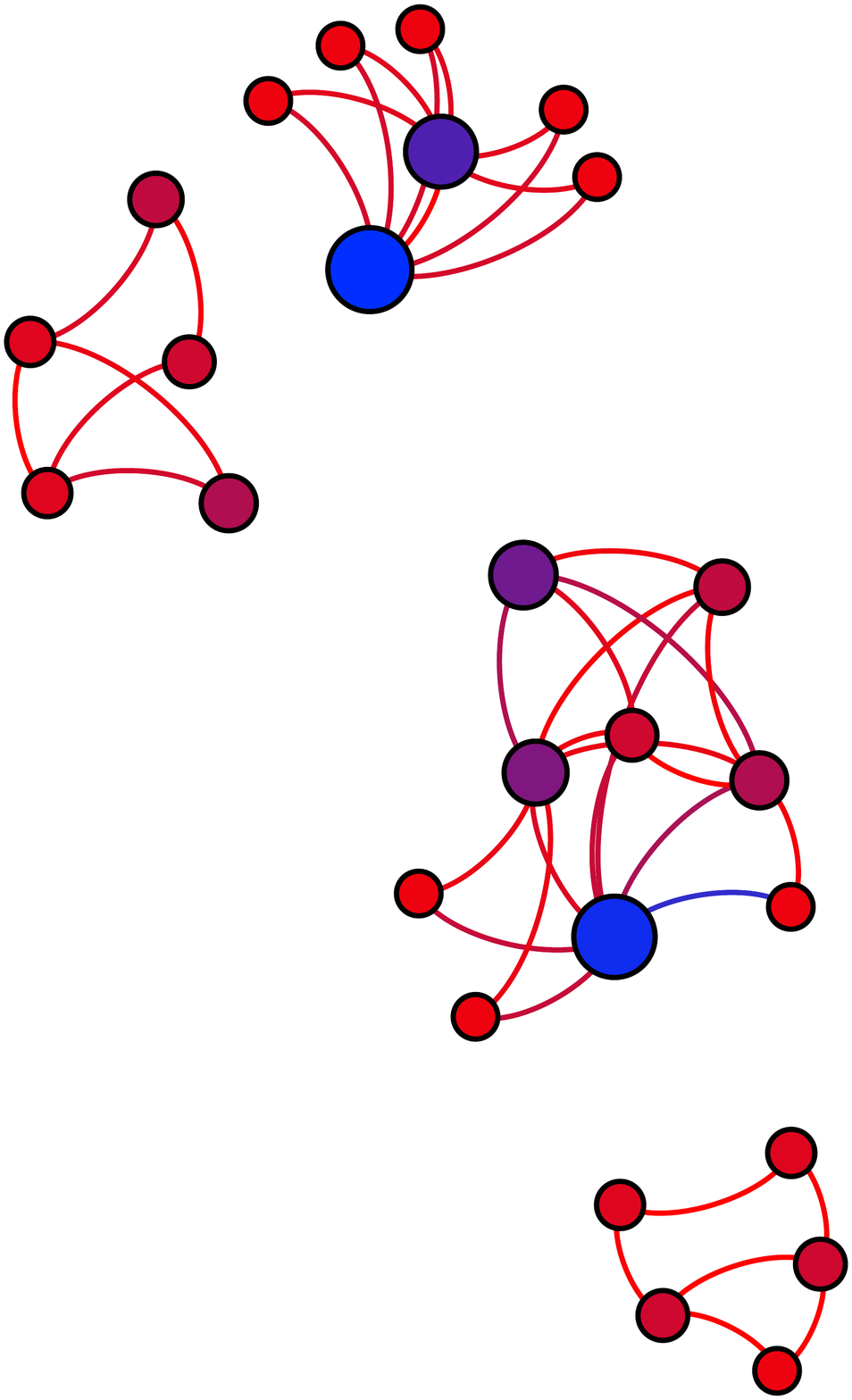}
}%
\centering
\caption{This is an example of a network of Karate Club~\cite{girvan2002community}. We use Ricci curvature as edge weights to represent different local structures (\textcolor{blue}{\textbf{tree-like}} and \textcolor{red}{\textbf{cyclic}} structures). It shows that one network contains heterogeneous topologies. }
\label{fig:example}
\end{figure}

Many generative adversarial networks (GAN) based methods~\cite{wang2018graphgan,dai2018adversarial,yu2018learning,zhu2021adversarial,li2021robust} have been proposed to solve the above problem by using adversarial training regularization. 
However, there are two limitations of the existing GAN-based methods for robust representation of real-world graphs: 

\textbf{Discrete topology representation.} 
Since the network is non-Euclidean geometry, the implementation of GAN-based methods on the network usually requires the topological-preserving constraints and performs the discrete operations in the network such as walking and sampling. 
Although these GAN-based methods can learn robust node representations, their generators focus on learning the discrete node connection distribution in the original graph.
The generated nodes do not accurately capture the graph topology and lead to serious distortion for graph representation. 
To address this issue, we aim to find a way to deal with the discrete topology of graphs as if it were continuous data. 
Fortunately, \textit{Riemannian geometry}~\cite{willmore2013introduction} provides a solid and manipulable mathematical framework. 
In recent works, certain types of graph data (e.g. hierarchical, scale-free, or cyclical graphs) have been shown to be better represented in non-Euclidean (Riemannian) geometries ~\cite{NickelK17Poincare,gu2019learning,bronstein2017geometric,defferrard2019deepsphere,sun2020perfect,sun2021hyperbolic,sun2021self,fu2021ace}.
Inspired by the unsupervised manifold hypothesis~\cite{cayton2005algorithms,narayanan2010sample,rifai2011manifold}, we can understand a graph as a discretization of a latent geometric manifold\footnote{In this paper, the manifold is equivalent to embedding space, and we will use the terms \textit{manifold} and \textit{space} interchangeably.}. 
For example, the hyperbolic geometric spaces (with negative curvature) can be intuitively understood as a continuous tree~\cite{Krioukov2010Hyperbolic,bachmann2020constant} and spherical geometry spaces (with positive curvature) benefit for modeling cyclical graphs~\cite{defferrard2019deepsphere,davidson2018hyperspherical,xu2018spherical,gu2019learning}. 
In these cases, the Riemannian geometric manifold has significant advantages by providing a better geometric prior and inductive bias for graphs with respective topological properties.
Inspired by this property, we propose to learn robust graph representation in Riemannian geometric spaces.

\begin{figure*}[ht]
\centering
\subfigure[Hyperbolic geometric space ($\kappa<0$).]{
\includegraphics[width=0.33\linewidth]{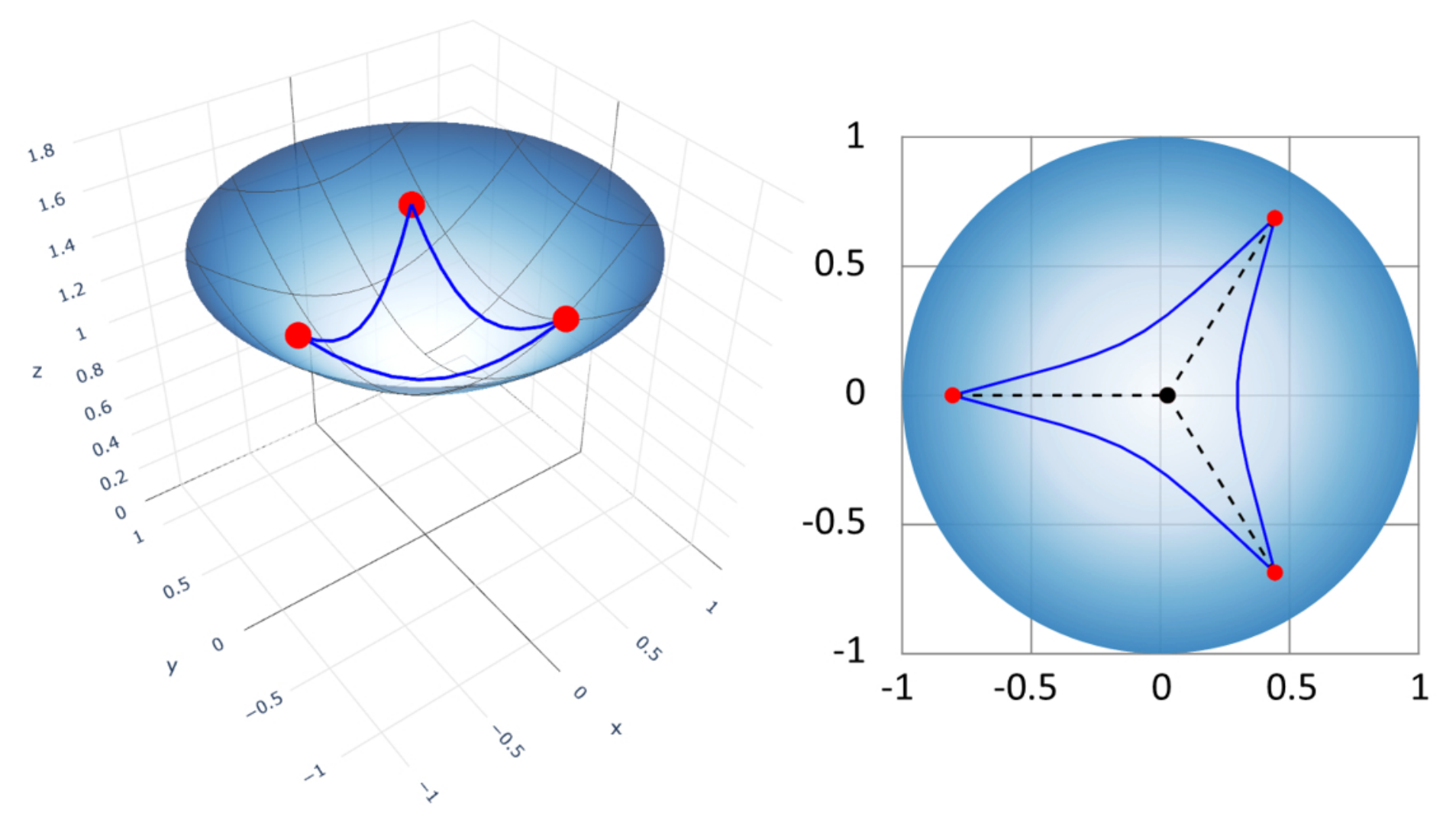}
}%
\subfigure[Euclidean space ($\kappa=0$).]{
\includegraphics[width=0.33\linewidth]{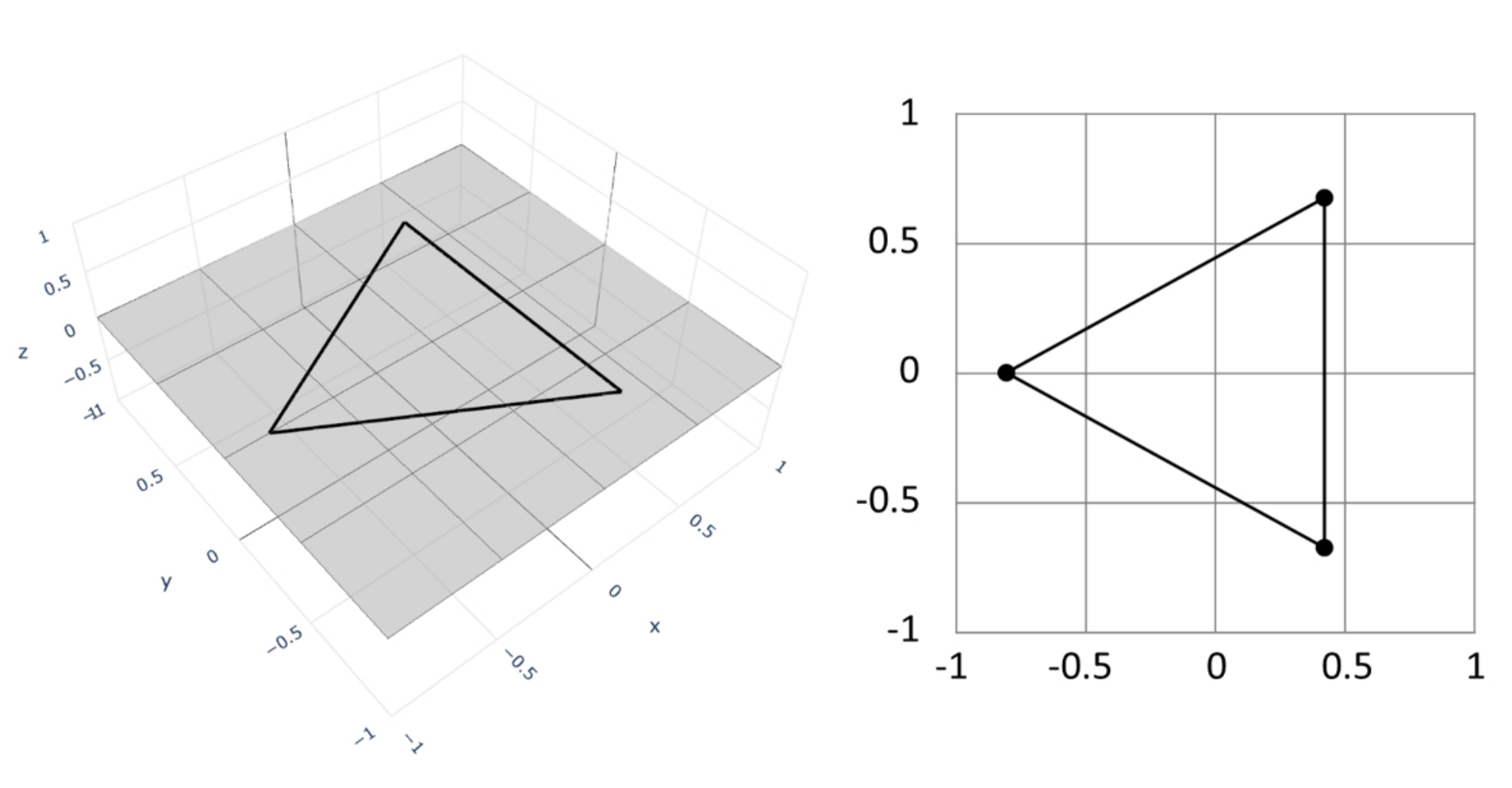}
}%
\subfigure[Spherical geometric space ($\kappa>0$).]{
\includegraphics[width=0.33\linewidth]{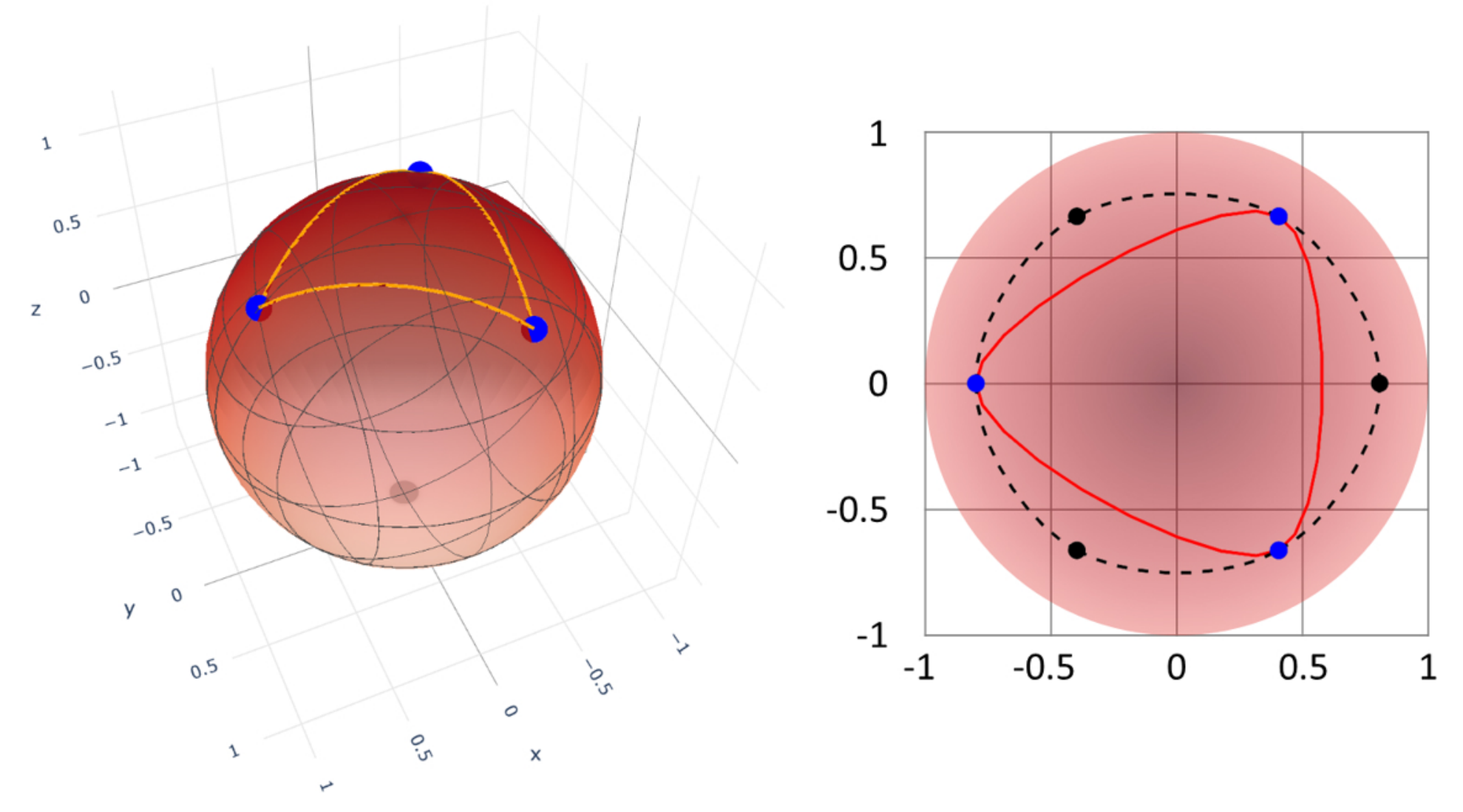}
}%
\centering
\caption{A toy example of a triangular structure in a Riemannian geometric space with different average sectional curvatures. (a) The geodesics (\textcolor{blue}{\textbf{blue solid lines}}) of a triangular structure approximate a tree (\textbf{dash lines and the virtual center}) in hyperbolic geometric space. 
(b) A triangular structure in Euclidean space. 
(c) The triangular geodesics (\textcolor{red}{\textbf{red solid lines}}) approximate a cycle structure (\textbf{dash lines and black virtual nodes}) in spherical geometric space. }
\label{fig:geometries}
\end{figure*}

\textbf{Topological heterogeneity.} 
There are some important properties (e.g., scale-free and small-world) usually presented by tree-like and cyclic structures~\cite{Krioukov2010Hyperbolic,papadopoulos2012popularity}. 
Meanwhile, these topological properties are also reflected in the density of the graph structure. 
As shown in Figure~\ref{fig:example}, the real-world graph data commonly has local structures with different topological properties, i.e., heterogeneous topologies. 
Moreover, the noisy edges~\cite{sun2021graph} may seriously change the topological properties of local structures (e.g. a single binary tree structure with an extra edge may become a triangle). 
However, the Riemannian geometric manifold with constant curvature is regarded as global geometric priors of graph topologies, and it is difficult to capture the topological properties of local structures. 

To address the above problems, we propose a novel \underline{\textbf{Curv}}ature Graph \underline{\textbf{G}}enerative \underline{\textbf{A}}dversarial \underline{\textbf{N}}etworks (\textbf{\modelname}). 
Specifically, we use a \textit{constant curvature}\footnote{In this paper, the constant curvature of the graph is considered as the global average sectional curvature. } to measure the global geometric prior of the graph, and generalize GAN into the Riemannian geometric space with the constant curvature. 
As shown in Figure~\ref{fig:geometries}, an appropriate Riemannian geometric space ensures minimal topological distortion for graph embedding, leading to better and robust topology representation. 
For the discrete topology global representation, \modelname~can directly perform any operations of GAN in the continuous Riemannian geometric space. 
For the topological heterogeneity issue, we design the \textit{Ricci curvature regularization} to improve the local structures capture capability of the model.
Overall, the contributions are summarized as follows: 
\begin{itemize}[leftmargin=*]
\item 
We propose a novel Curvature Graph Generative Adversarial Networks (\modelname), which is the first attempt to learn the robust node representations in a unified Riemannian geometric space. 
\item 
\modelname~ can directly generate fake neighbor nodes in continuous Riemannian geometric space conforming to the graph topological prior and better preserve the global and local topological proprieties by introducing various curvature metrics. 
\item Extensive experiments on synthetic and real-world datasets demonstrate a significant and consistent improvement in model robustness and efficiency with competitive performance. 
\end{itemize} 

\section{Preliminary}
\subsection{Riemannian Geometric Manifold}
Riemannian geometry is a strong and elegant mathematical framework for solving non-Euclidean geometric problems in machine learning and manifold learning~\cite{bronstein2017geometric,defferrard2019deepsphere,lin2008riemannian}. 
A manifold is a special kind of connectivity in which the coordinate transformation between any two (local) coordinate systems is continuous. 
\textit{Riemannian manifold}~\cite{alexander1978michael} is a \textit{smooth manifold} $\mathcal{M}$ of dimension $d$ with the \textit{Riemannian metric} $g$, denoted as $(\mathcal{M}, g)$. 
At each point $\mathrm{x} \in \mathcal{M}$, the locally space looks like a $d$-dimension space, and it associates with the \textit{tangent space} $\mathcal{T}_{\mathrm{x}}\mathcal{M}$ of $d$-dimension. 
For each point $\mathrm{x}$, the Riemannian metric $g$ is given by an inner-product $g_{\mathrm{x}}(\cdot,\cdot): \mathcal{T}_{\mathrm{x}}\mathcal{M} \times \mathcal{T}_{\mathrm{x}}\mathcal{M} \to \mathbb{R}$. 


\subsection{The $\kappa$-stereographic Model}
The gyrovector space~\cite{ungar1999hyperbolic,ungar2005analytic,ungar2008gyrovector,ungar2014analytic} formalism is used to generalize vector spaces to the Poincar\'e model of hyperbolic geometry~\cite{PoincareGlove}. 
The important quantities from Riemannian geometry can be rewritten in terms of the M\"obius vector addition and scalar-vector multiplication~\cite{HNN:GaneaBH18}. 
However, these mathematical tools are used only in hyperbolic spaces (i.e., constant negative curvature).
To extend these tools to unify all curvatures,~\cite{bachmann2020constant} leverage gyrovector spaces to the $\kappa$-stereographic model.

Given a curvature $\kappa \in \mathbb{R}$, a $n$-dimension $\kappa$-\textit{stereographic model} $(\mathfrak{st}_{\kappa}^{d}, g^{\kappa})$ can be defined by the manifold $\mathfrak{st}_{\kappa}^{d}$ and Riemannian metric $g^{\kappa}$:
\begin{align}
\mathfrak{st}_{\kappa}^{d} &= \{ \mathbf{x} \in \mathbb{R}^{d} | - \kappa \| \mathbf{x} \|_{2}^{2} < 1 \}, \\
g^{\kappa}_{\mathbf{x}} &= {\lambda^{\kappa}_{\mathbf{x}}}^{2} g^{\mathbb{E}}, \  \mathrm{where} \  \lambda^{\kappa}_{\mathbf{x}} = 2/(1+\kappa\|x\|^{2}),
\end{align}
where $g^{\mathbb{E}} = \mathbf{I}^{d}$ is the Euclidean metric tensor. 
Note that when the curvature $\kappa > 0$, $\mathfrak{st}_{\kappa}^{d}$ is $\mathbb{R}^{d}$, while for $\kappa < 0$, $\mathfrak{st}_{\kappa}^{d}$ is a Poincar\'e ball of radius $1/\sqrt{-\kappa}$. 

\textbf{Distance.}
For any point pair $\mathbf{x}, \mathbf{y} \in \mathfrak{st}_{\kappa}^{d}$, $\mathbf{x} \ne \mathbf{y}$, the projection node $\mathbf{v} \ne 0$ the distance in $\kappa$-stereographic space is defined as:
\begin{align}
\mathbf{d}^{\kappa}_{\mathfrak{st}}(\mathbf{x}, \mathbf{y})=(2 / \sqrt{|\kappa|}) \tan _{\kappa}^{-1}\left\|-\mathbf{x} \oplus_{\kappa} \mathbf{y}\right\|. 
\label{Eq:distance}
\end{align}
Note that the distance is defined in all the cases except for $\mathbf{x} \ne -\mathbf{y}/\kappa \| \mathbf{y} \|^{2} \  \mathrm{if} \  \kappa > 0$. 

\textbf{Exponential and Logarithmic Maps.}
The manifold $\mathfrak{st}_\kappa^d$ and the tangent space $\mathcal{T}_{\mathrm{x}}\mathfrak{st}$ can be mapped to each other via \textit{exponential map} and \textit{logarithmic map}. 
The \textit{exponential map} $\mathrm{exp}_{\mathbf{x}}^{\kappa}(\cdot)$ and \textit{logarithmic map} $\log _{\mathbf{x}}^{\kappa}(\cdot)$ are defined as: 
\begin{align}
\small
\mathrm{exp}_{\mathbf{x}}^{\kappa}(\mathbf{v}) &:= \mathbf{x} \oplus_{\kappa} \left (\tan_{\kappa} \left( \sqrt{|\kappa|} \frac{\lambda_{\mathbf{x}}^{\kappa} \left \| \mathbf{v} \right \|}{2} \right ) \frac{\mathbf{v}}{\| \mathbf{v} \|} \right), \\
\log _{\mathbf{x}}^{\kappa}(\mathbf{y}) &:= \frac{2|\kappa|^{-\frac{1}{2}}}{\lambda_{\mathbf{x}}^{\kappa}} \tan _{\kappa}^{-1}\left\|-\mathbf{x} \oplus_{\kappa} \mathbf{y}\right\| \frac{-\mathbf{x} \oplus_{\kappa} \mathbf{y}}{\left\|-\mathbf{x} \oplus_{k} \mathbf{y}\right\|},
\end{align}
where $\tan _{\kappa} = \frac{1}{\sqrt{|\kappa}|}\tan (\cdot)$ and $\lambda_{\mathbf{x}^{\kappa}}$ is the conformal factor which comes from Riemannian metric. 

\subsection{The Graph Curvature}
\textbf{Sectional Curvature.}
In Riemannian geometry, the sectional curvature~\cite{ni2015ricci} is one of the ways to describe the curvature of Riemannian manifolds. 
In existing works~\cite{bachmann2020constant,gu2019learning,skopek2020mixedcurvature}, average sectional curvature $\kappa$ has been used as the constant curvature of non-Euclidean geometric embedding space. 

\noindent\textbf{Ricci Curvature.}
Ricci curvature~\cite{lin2011ricci,ni2015ricci} is a broadly metric which measures the geometry of a given metric tensor that differs locally from that of ordinary Euclidean space. 
In machine learning, Ricci curvature is transformed into edge weights to measure local structural properties~\cite{ye2019curvature}. 
Ollivier-Ricci curvature~\cite{ollivier2009ricci} is a coarse approach used to compute the Ricci curvature for discrete graphs. 

\section{\modelname~Model}
\begin{figure*}[htb]
\centering
\includegraphics[width=0.96\textwidth]{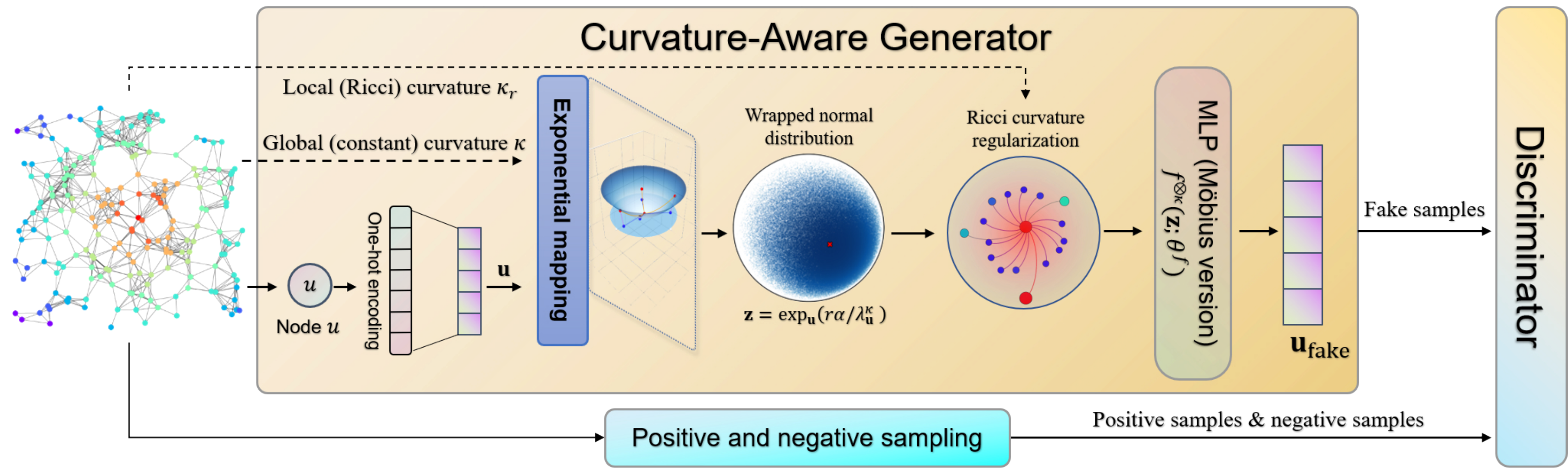}
\caption{\textbf{An illustration of \modelname~ architecture.} 
(1) \modelname~ estimates the constant curvature $\kappa$ and Ricci curvature $\kappa_r$ of the graph; 
(2) \modelname~'s generator generates negative samples by sampling fake node representations from the wrapper normal distribution; 
(3) \modelname~'s discriminator discriminates the positive, negative and generated fake samples. }
\label{fig:Architecture}
\end{figure*}

In this section, we present a novel Curvature Graph Generative Adversarial Network (\modelname) in the latent geometric space of the graph. 
The overall architecture is shown in Figure~\ref{fig:Architecture}. 

\subsection{Geometric Prior of Graph Topology}
An interesting theory of graph geometry is that some typical topological structures can be described intuitively using Riemannian geometry with different curvature $\kappa$~\cite{networkgeo}, i.e., hyperbolic ($\kappa<0$), Euclidean ($\kappa=0$) and spherical ($\kappa>0$) geometries. 
As shown in Figure~\ref{fig:geometries}, the hyperbolic space can be intuitively understood as a continuous tree~\cite{Krioukov2010Hyperbolic}, and spherical geometry provides benefits for learning cyclical structure~\cite{wilson2014spherical,xu2018spherical,grattarola2018learning,davidson2018hyperspherical, gu2019learning}. 
Therefore, we can learn better graph representation with minimal embedding distortion in an appropriate Riemannian geometric space~\cite{bachmann2020constant}. 
Motivated by this idea, we first search for an appropriate Riemannian geometric space to approximate the global topological properties of the graph, and then we capture the local structural features for each node by introducing Ricci curvature. 
In this way, we propose a curvature-constrained framework to capture both the global topology and local structure of the graph. 


\textbf{Global Curvature Estimation.}
In machine learning, the Riemannian manifold is commonly considered as a geometric prior with constant curvature. 
For a graph embedding in Riemannian geometric space, a key parameter is \textit{constant curvature} $\kappa$, which can affect the embedding distortion of a graph topology~\cite{gu2019learning}. 
To minimize the embedding distortion and explore the optimal curvature, we leverage the average sectional curvature estimation algorithm~\cite{gu2019learning,bachmann2020constant} to estimate global curvature. 
Specifically,let $(a,b,c)$ be a geodesic triangle in manifold $\mathfrak{st}^{d}_{\kappa}$, and $m$ be the (geodesic) midpoint of $(b,c)$.
Their quantities are defined as follows: 
\begin{equation}\label{curvest}
\begin{aligned}
    &\xi_{\mathfrak{st}}(a,b;c)\! =\! p_{\mathfrak{st}}\!(a,m)^2\! +\! \frac{p_{\mathfrak{st}}\!(b,c)^2}{4}\! +\! \frac{p_{\mathfrak{st}}(a,b)^{2}\!+\!p_{\mathfrak{st}}\!(a,c)^2}{2}, \\
    &\xi_{\mathfrak{st}}(m;a,b;c) = \frac{1}{2 p_{\mathfrak{st}}(a,m)} \xi_{\mathfrak{st}}(a,b;c).
\end{aligned}
\end{equation}

We design our curvature updating according to Eq.~\eqref{curvest}. 
The new average curvature estimation $\kappa$ is defined as: 
\begin{equation}
\label{Eq:curvature}
\begin{aligned}
    \kappa &= \frac{1}{|V|}\sum_{m\in V}\left (\frac{1}{n_s}\sum_{j=0}^{n_s}\xi_{\mathfrak{st}}\left (\mathbf{h}_{m};\mathbf{h}_{a_j},\mathbf{h}_{b_j};\mathbf{h}_{c_j}\right )\right ), 
\end{aligned}
\end{equation}
where $b$ and $c$ are randomly sampled from the neighbors of $m$, and $a$ is a node in the graph $G$ except for $\{m, b, c\}$. 
For each node, we sample $n_s$ times and take the average as the estimated curvature. 

\textbf{Local Curvature Computation.}
To deal with the embedding distortion caused by topological heterogeneity, we also need to consider the local structural properties of the graph.
We leverage the \textit{Ollivier-Ricci curvature}~\cite{ollivier2009ricci} to solve this problem.
Specifically, Ollivier-Ricci curvature $\kappa_{r}(x,y)$ of the edge $(x,y)$ is defined as:
\begin{equation}
\label{Eq:ricci-curvature}
\begin{aligned}
    \kappa_{r}(x,y) &= \frac{W(m_x,m_y)}{d(x,y)}, 
\end{aligned}
\end{equation}
where $W(\cdot,\cdot)$ is the Wasserstein distance, $d(\cdot,\cdot)$ is the geodesic distance (embedding distance), and $m_x$ is the mass distribution of node $x$. 
The mass distribution represents the importance distribution of a node and its one-hop neighborhood~\cite{ollivier2009ricci}, which is defined as:

\begin{equation}
\label{Eq:mass_dist}
\begin{aligned}
    m^{\alpha}_x =
    \begin{cases}
    \alpha & \mathbf{ if } x_i = x,\\
    1-\alpha & \mathbf{ if } x_i \in \mathrm{Neighbor}(x)\\
    0 & \mathbf{ otherwise, }
    \end{cases}, 
\end{aligned}
\end{equation}
where $\alpha$ is a hyper-parameter that represents the importance of node $x$ and we take $\alpha = 0.5$ following the existing works~\cite{sia2019ollivier, ye2019curvature}.

\subsection{Curvature-Aware Generator}
Curvature-Aware Generator aims to directly generate fake nodes from the continuous Riemannian geometric space of the global topology as the enhanced negative samples while preserving the local structure information of the graph as much as possible. 

\textbf{Global Topological Representation Generation.}
First, we map a node $u$ into the Riemannian geometric space with the global curvature $\kappa$, and input it to an embedding layer of M\"obius version to get a dense vector representation. 
In order to generate appropriate representations of fake neighbor nodes, we need to extend the noise distribution to the Riemannian geometric space. 
An approach is to use exponential mapping to map a normal distribution into a Riemannian manifold, and this distribution measure is referred to as the \textit{wrapped normal distribution}~\cite{grattarola2019adversarial,nagano2019differentiable}.
The $d$-dimensional wrapped normal distribution $\mathcal{N}^{\kappa}_{\mathfrak{st}} (\mathbf{z} | \boldsymbol{\mu},\sigma^{2} \mathrm{I})$ is defined as:
\begin{align}
\label{Eq:WrappedNormal}
\small
&\mathcal{N}^{\kappa}_{\mathfrak{st}} (\mathbf{z} | \boldsymbol{\mu},\sigma^{2} \mathrm{I})  \\ 
&=\mathcal{N}\left(\lambda_{\mu}^{\kappa} \log _{\mu}(\boldsymbol{z}) \mid \mathbf{0}, \sigma^{2} \mathrm{I}\right)\left(\frac{\sqrt{\kappa} \mathbf{d}_{\mathfrak{st}}^{\kappa}(\boldsymbol{\mu}, \boldsymbol{z})}{\sinh \left(\sqrt{\kappa} \mathbf{d}_{\mathfrak{st}}^{\kappa}(\boldsymbol{\mu}, \boldsymbol{z})\right)}\right)^{d-1}, \nonumber
\end{align}
where $\mu \in \mathfrak{st}^{d}_{\kappa}$ is the mean parameter, and $\sigma \in \mathbb{R}^{d}$ is the variance. 
Then we can introduce a reparameterized sampling strategy to generate the negative node representation. 
Specifically, for a fake node $\mathbf{u}_\mathrm{fake}$ in the manifold $\mathfrak{st}$, we can generate the noisy representation in the Riemannian geometric space, given by
\begin{align}
\small
\mathbf{z}^{\mathcal{G}}_{\mathbf{u}_\mathrm{fake}} = \mathrm{exp}^{\kappa}_{\textbf{u}} \left( \frac{r}{\lambda^{\kappa}_{\mathbf{u}}} \alpha \right), \mathrm{with} \ \mathbf{u}_\mathrm{fake} \sim \mathcal{N}(0, \sigma^{2} \mathrm{I}),
\label{Eq:sample}
\end{align}
with radius $r = \mathbf{d}^{\kappa}_{\mathfrak{st}}$, direction $\alpha = \frac{\mathbf{v}}{r}$, and $\mathbf{u} \in \mathfrak{st}^{d}_{\kappa}$ is the embedding vector of node $u$. 

\textbf{Local Structure Preserving Regularization.}
Since the real-world graph data usually has a heterogeneous topology, the edge generated by data noise may seriously change the local structure, leading to the following two issues: 
(1) the model cannot capture the correct local structure, leading to the generated sample may have different properties from the original structure:
(2) the fake samples generated by the model are no longer indistinguishable from the real samples. 
Therefore, we need to regularize the generated fake nodes by using a measurement of preserving local structure information. 
If two nodes are similar in the same local structure, they should have the same Ricci curvatures in the neighborhood.

According to Eq.~\eqref{Eq:ricci-curvature}, the regularization term can given by
\begin{equation}
\begin{aligned}
\small
    \textrm{Reg}(\kappa_r^u,\kappa_r^{u_\textrm{fake}}) = \frac{1}{|\mathbb{N}(u)|}\! \sum_{v \in \mathbb{N}(u)} \! \left (1- \frac{W(m_u,m_v)d(u_\textrm{fake},v)}{W(m_{u_\textrm{fake}},m_v)d(u,v)} \right ),
\label{Eq:regularization}
\end{aligned}
\end{equation}
where $\mathbb{N}(u)$ is the neighbors of node $u$. To facilitate the calculation of Ricci curvature~\cite{ollivier2009ricci}, we assume the generated fake node has a similar one-hop neighborhood as the original node $u$, i.e., $m_u = m_{u_\textrm{fake}}$. 

\textbf{Fake Sample Generation Strategy.}
Given a node $u$, the generator outputs the embedding $\mathbf{z}^{\mathcal{G}}_{\mathbf{u}_\mathrm{fake}} = G^{\kappa}_{\mathfrak{st}}\left(\mathbf{u}; \theta^{G^{\kappa}_{\mathfrak{st}}}\right)$ of the fake node $\mathbf{u}_{\mathrm{fake}}$ as a substitute for node $\mathbf{u}$. 
In this way, we can obtain a node-pair set $\{(v, u_{\mathrm{fake}}), v \in \mathcal{N}(u)\}$ as the negative samples. 
The generator of Riemannian latent geometry is defined as: 
\begin{equation}
\begin{aligned}
\small
G^{\kappa}_{\mathfrak{st}} \left ( \cdot; \theta^{G^{\kappa}_{\mathfrak{st}}} \right ) &= f^{\otimes_{\kappa}} \left (\mathbf{z}; \theta^{f}\right),\\
f^{\otimes_{\kappa}} &= \mathrm{exp}^{\kappa}_{\mathbf{u}} \left(f\left(\mathrm{log}^{\kappa}_{\mathbf{u}}\left(\mathbf{u}\right)\right)\right),
\label{Eq:generator}
\end{aligned}
\end{equation}
where $\theta ^{G^{\kappa}_{\mathfrak{st}}}$ is the parameters for generator $G^{\kappa}_{\mathfrak{st}}$, and $f^{\otimes_{\kappa}}$ is a M\"obius version of the multi-layer perception.  

\textbf{Optimization of Generator.}
For the high-order proximity of each node-pair in a graph, the advantage of our CurvGAN is that it doesn't require traversing the shortest path between two points to compute the connectivity probability. 
According to Eq.~\eqref{Eq:regularization}, when the embedding distortion is minimal, the longer the shortest path between any two nodes, the less probability of this node-pair by direct calculation in the latent geometric space, and vice versa. 
The loss function of the generator is defined as follows:
\begin{align}
\small
\mathcal{L}_G = &\mathbb{E}_{u \in \mathcal{V}} \mathrm{log}\left(1-D^{\kappa}\left(\mathbf{u}^{D},\mathbf{u}_{\mathrm{fake}}\right)\right) \nonumber \\
&+ \lambda \left \| \sum_{u \in V} \textrm{Reg}(\kappa_r^u,\kappa_r^{u_\textrm{fake}})  \right \|_2. 
\label{Eq:loss_generator}
\end{align}


\subsection{Discriminator}
The discriminator of \modelname~aims to determine whether the connection between two nodes is real. 
For node pair $(u, v)$, the discriminant function outputs the connection probability between two nodes. 
In general, the discriminator $D$ is defined as:
\begin{align} 
\small
D^{\kappa}(u,v;\theta^{D})=\mathcal{F}_{\kappa}\left(u, v ; \theta^{F}\right),
\label{eq4}
\end{align}
where $\theta^{D}$ is the parameters of discriminator $D$. 

For \textit{link prediction} task, we use the Fermi-Dirac decoder~\cite{Krioukov2010Hyperbolic,NickelK17Poincare} to compute the connection probability: 
\begin{equation}
\begin{aligned}
    \mathcal{F}_{\kappa}\left(u, v ; \theta^{\mathcal{F}}\right) = \left [ e^{ (\mathbf{d}^{\kappa}_{\mathfrak{st}}(u,v)^2 - \rho) / \tau} + 1 \right ]^{-1}, 
\end{aligned}
\end{equation}
where $\mathbf{d}^{\kappa}_{\mathfrak{st}}$ is the hyperbolic distance in Eq.~\eqref{Eq:distance} and $(\rho$, $\tau)$ are hyper-parameters of Fermi-Dirac distribution. 


For \textit{node classification} task, inspired by~\cite{HGCN_ChamiYRL19}, we map the output embedding $\textbf{z}_i,i \in \mathcal{V}$ to the tangent space $\mathcal{T}_{\mathrm{o}}\mathfrak{st}$ by logarithmic mapping $\log ^{\kappa} _{\mathbf{o}}(z_i)$, then perform Euclidean multinomial logistic regression, where $\mathrm{o}$ is the north pole or origin. 

For a graph $\mathcal{G}(\mathcal{V},\mathcal{E})$, the input samples of the discriminator are as follows:
\begin{itemize}[leftmargin=*]
\item Positive Sample $(u, v)$: 
There indeed exists a directed edge from $u$ to $v$ on a graph $\mathcal{G}$. 
\item Negative Samples $(u,w)$ and $(u_{\mathrm{fake}},v)$:
For a given node $u,w,v \in \mathcal{V}$, the negative samples consists of the samples by negative sampling in the original graph $(u,w) \notin \mathcal{E}$ and the fake samples $(u_{\mathrm{fake}},v)$ generated by the generator. 
\end{itemize}


\begin{algorithm}[t]
    \LinesNumbered
    \caption{\modelname} 
    \label{Alg:MAGRL}
    \KwIn{Graph $\mathcal{G} = \{V, E\}$; Number of training epochs $n_e$, generator’s epochs $n^G_e$, discriminator’s epochs $n^D_e$; Number of samples $n_s$.}
    \KwOut{Predicted result of the downstream task.}
    \tcp{Curvature Estimation }
    $\kappa \gets$ Eq.~\eqref{Eq:curvature};   \\
    $\kappa_r \gets$ Eq.~\eqref{Eq:ricci-curvature};   \\
    \For{$t = 1,2,\cdots,n_e$}{
        \tcp{Train Discriminator}
        \For{$d = 1,2,\cdots,n^D_e$}{
            \tcp{Sample neighbour $v \in \mathcal{N}(u)$ for each $u \in V$}
            $\mathbf{u}_{\mathrm{pos}} \gets \textsc{RandomWalk}(u, n_s);$    \\ 
            \tcp{Sample negative nodes and generate fake nodes}
            $\mathbf{u}_{\mathrm{neg}} \gets \textsc{RandomSelect}(V, n_s);$   \\
            $\mathbf{u}_{\mathrm{fake}} \gets$ Eq.~\eqref{Eq:generator};  \\
            \tcp{Optimize}
            $\mathcal{L}_D \gets $ Eq.~\eqref{Eq:loss_generator};   \\
        }
        \tcp{Train Generator}
        \For{$g = 1,2,\cdots,n^G_e$}{
            \tcp{Generate fake nodes}
            $\mathbf{u}_{\mathrm{fake}} \gets $ Eq.~\eqref{Eq:generator}; \\
            \textrm{Reg}($\kappa^{\mathrm{fake}}_r, \kappa^{\mathrm{true}}_r) \gets$ Eq.~\eqref{Eq:regularization};\\
            \tcp{Optimize}
            $\mathcal{L}_G \gets$ Eq.~\eqref{Loss:disicriminator}; \\
        }
    }    
\end{algorithm}
\textbf{Optimization of Discriminator.}
The loss function of positive samples $(u, v)$ is:
\begin{align}
\mathcal{L}_{\rm pos}^{D} = \mathbb{E}_{(u, v) \sim \mathcal{E}} -\log D^\kappa (\mathbf{u}^D, \mathbf{u}^D_{\mathrm{pos}}). 
\label{eq7}
\end{align}
The loss function of negative samples $(u,w)$ and $(u_{\mathrm{fake}},v)$ is defined as: 
\begin{align}
\mathcal{L}_{\rm neg}^{D} =& \mathbb{E}_{u \in \mathcal{V}} -( \log (1-D^\kappa (\mathbf{u}^D, \mathbf{u}^D_{\mathrm{neg}})) \nonumber \\
& + \log (1 - D^\kappa (\mathbf{u}^D, \mathbf{u}_{\mathrm{fake}}))).
\label{eq8}
\end{align}
Then we integrate the above Eq.~\eqref{eq7} and Eq.~\eqref{eq8} as the loss function of the discriminator, which we try to minimize: 
\begin{align}
\mathcal{L}^{D} = \mathcal{L}_{\rm pos}^{D} + \mathcal{L}_{\rm neg}^{D}. 
\label{Loss:disicriminator}
\end{align}%

\subsection{Training and Complexity Analysis}
The overall training algorithm for \modelname~is summarized in Algorithm~\ref{Alg:MAGRL}. 
For each training epoch, time complexity of \modelname~ per epoch is $O \left(n_s \cdot (n^D_e \cdot |\mathcal{E}| + n^G_e \cdot (|\mathcal{V}|+n_s\mathrm{log}(n_s))) \cdot d^2 \right)$. 
Since $n_s$, $n^G_e$, $n^D_e$ and $d$ are small constants, \modelname~'s time complexity is linear to $|\mathcal{V}|$ and $|\mathcal{E}|$. 
The space complexity of \modelname~is $O(2 \cdot d \cdot (|\mathcal{V}| + |\mathcal{E}|))$. 
In conclusion, \modelname~is both time and space-efficient, making it scalable for large-scale graphs.

\begin{table}[t]
\small
\caption{Statistics of datasets.}
\centering
\begin{tabular}{cl|rrrrc}
\toprule
\multicolumn{2}{c|}{\textbf{Dataset}}  & \textbf{\#Nodes} & \textbf{\#Edges} &\textbf{Avg. Degree} & \textbf{\#Labels} & \textbf{$\kappa$} \\ 
\midrule
\multirow{3}{*}{\rotatebox{90}{\textbf{Synth.}}}  &\textbf{SBM}     & 1,000  & 15,691 & 15.69  & 5     & -1.496    \\
&\textbf{BA}  & 1,000 & 2,991 & 2.99  & 5    & -1.338     \\
&\textbf{WS}    & 1,000  & 11,000 & 11.00  & 5   & 0.872    \\
\midrule
\multirow{3}{*}{\rotatebox{90}{\textbf{Real}}}  &\textbf{Cora}     & 2,708  & 5,429 & 3.90  & 7     & -2.817    \\
&\textbf{Citeseer}  & 3,312 & 4,732 & 2.79  & 6    & -4.364     \\
&\textbf{Polblogs}    & 1,490  & 19,025 & 25.54  & 2   & -0.823    \\
\bottomrule
\end{tabular}
\label{dataset_description}
\end{table}

\section{Experiment}
\label{sec:Experiment}
In this section, we conduct comprehensive experiments to demonstrate the effectiveness and adaptability of \modelname~\footnote{Code is available at \url{https://github.com/RingBDStack/CurvGAN}.} on various datasets and tasks. 
We further analyze the robustness to investigate the expressiveness of \modelname. 

\begin{table*}[htbp]
\caption{Summary of link prediction AUC scores (\%), node classification Micro-F1 and Macro-F1 scores (\%) on synthetic graphs.\\
(Result: average score ± standard deviation; \textbf{Bold}: best; \underline{Underline}: runner-up.)}
\vspace{-1em}
\centering
\resizebox{\textwidth}{!}{
\begin{tabular}{c|ccc|ccc|ccc|c}
\toprule
\multirow{2}{*}{\textbf{Method}} & \multicolumn{3}{c}{\textbf{Stochastic Block Model}}    & \multicolumn{3}{c}{\textbf{Barabási-Albert}}    & \multicolumn{3}{c|}{\textbf{Watts-Strogatz}}    & \multirow{2}{*}{\textbf{Avg.Rank}}\\ \cline{2-10} 
                        & AUC                & Micro-F1               & Macro-F1             & AUC               & Micro-F1                & Macro-F1              & AUC               & Micro-F1                & Macro-F1               \\ 
\midrule
GAE~\cite{kipf2016variational}    & 50.13±0.12 & 24.07±1.12 & 21.94±2.11        & 50.26±0.21   & 39.35±1.72  & 17.83±1.11        & 50.10±0.08   & 19.08±1.87   & 16.47±2.64  & 6.6  \\ 
VGAE~\cite{kipf2016variational}   & 50.32±1.49 & 20.47±2.05 & 15.41±1.11    & 62.43±1.26   & 37.44±1.73  & 15.88±2.31    & 49.94±0.57   & 19.14±1.40   & 12.02±1.13  & 7.2  \\ 
DGI~\cite{Velickovic2019DeepGI}    & 49.88±0.51 & 19.06±1.87 & 12.13±2.08        & 70.90±2.12   & 38.24±1.11  & 18.13±0.26        & 49.55±0.49   & 18.27±1.14   & 13.31±0.80   & 8.1  \\ 
G2G~\cite{Bojchevski2018DeepGE}    & 79.45±1.28 & 21.44±0.05 & 20.98±0.03        & 54.29±1.62   & 42.27±0.31  & 23.96±0.39        & \underline{73.15±2.23}   & 22.89±0.07   & 22.75±0.07  & 5.1  \\ 
\midrule
GraphGAN~\cite{wang2018graphgan}    & 84.56±2.84 & 38.60±0.51 & 38.87±0.32        & 63.34±4.19  & 43.60±0.61  & \underline{24.57±0.53}        & 66.63±9.46   & \underline{41.80±0.84}   & \textbf{41.76±1.25}  & \underline{3.0}  \\ 
ANE~\cite{dai2018adversarial}    & 85.09±1.12 & 39.88±1.06 & 33.85±1.75        & 62.13±2.49   & \underline{46.04±3.01}  & 19.32±2.66        & 62.98±1.44   & 33.84±2.75   & 33.51±2.00 & 3.7 \\ 
\midrule
$\mathcal{P}$-VAE~\cite{mathieu2019poincare}  & \underline{86.10±0.97} & \underline{57.94±1.29} & \underline{52.97±1.47}        &\underline{76.08±1.22}    & 38.38±1.37  & 20.03±0.32        & 51.43±3.56   & 19.85±1.40   & 13.62±1.62   & 4.7  \\ 
Hype-ANE~\cite{HyperANE}    & 82.29±2.70 & 18.84±0.32 & 11.93±0.09        & 70.92±0.43   & \textbf{56.92±2.41}  & \textbf{31.58±1.17}        & 63.34±4.19   & 33.40±2.55   & 32.94±2.70     & 4.4 \\ 
\midrule
\textbf{\modelname~(Ours)}        & \textbf{89.74±0.70} & \textbf{59.00±0.56} & \textbf{55.99±2.20}        & \textbf{95.87±0.86}   & 42.50±1.37  & 19.28±0.64        & \textbf{88.67±0.22}   & \textbf{43.10±1.78}   & \underline{35.21±1.95}  & \textbf{1.8}  \\ 
\bottomrule
\end{tabular}}
\label{table:synthetic}
\end{table*}

\begin{table*}[htbp]
\caption{Summary of link prediction AUC scores (\%), node classification Micro-F1 and Macro-F1 scores (\%) on real-world graphs.\\
\vspace{-1em}
(Result: average score ± standard deviation; \textbf{Bold}: best; \underline{Underline}: runner-up.)}
\centering
\resizebox{\textwidth}{!}{
\begin{tabular}{c|ccc|ccc|ccc|c}
\toprule
\multirow{2}{*}{\textbf{Method}} & \multicolumn{3}{c}{\textbf{Cora}}    & \multicolumn{3}{c}{\textbf{Citeseer}}     &\multicolumn{3}{c|}{\textbf{Polblogs}}     & \multirow{2}{*}{\textbf{Avg.Rank}}     \\ 
\cline{2-10} 
                        & AUC                & Micro-F1               & Macro-F1             & AUC               & Micro-F1                & Macro-F1              & AUC               & Micro-F1                & Macro-F1               \\ 
\midrule
GAE~\cite{kipf2016variational}   & 86.12±0.87 & 80.92±0.99 & 79.55±1.32        & 87.25±1.26   & 58.50±3.31  & 50.41±3.32        & 83.55±0.62   & 89.50±0.53   & 89.42±0.53  & 4.4  \\ 
VGAE~\cite{kipf2016variational}   & 85.94±0.05 & 79.95±0.95 & 78.79±0.97   & 85.72±2.20   & 63.75±1.39  & 55.47±1.34  & 88.12±0.64   & 87.02±1.04    & 86.98±0.88   & 5.8    \\
DGI~\cite{Velickovic2019DeepGI}    & 75.39±0.29 & 74.09±1.75 & 66.70±1.91   & 81.30±3.57   & \textbf{73.16±0.68}   & \underline{63.27±0.63}  & 76.33±3.35   & 87.11±1.18    & 87.06±1.19 & 6.3  \\ 
G2G~\cite{Bojchevski2018DeepGE}    & 84.47±0.70 & 82.13±0.58 & 81.14±0.40   & \underline{90.34±1.44}   & \underline{71.03±0.27}  & \textbf{66.44±0.32}  & \underline{91.02±0.29}   & 87.52±0.28
    & 87.51±0.28    & \underline{3.3}    \\
\midrule
GraphGAN~\cite{wang2018graphgan}    & 82.50±0.64   & 76.40±0.21   & 76.80±0.34      & 74.50±0.02   & 49.80±1.02   & 45.70±0.13     & 69.80±0.26   & 77.45±0.64   & 76.90±0.43     & 8.4  \\
ANE~\cite{dai2018adversarial}    & 83.10±0.57   & \underline{83.00±0.51}   & \underline{81.90±1.40}          & 83.00±1.20   & 50.20±0.12   & 49.50±0.61          & 73.09±0.76   & \underline{95.07±0.65}    & \underline{95.06±0.65}   & 4.8 \\ 
\midrule
$\mathcal{P}$-VAE~\cite{mathieu2019poincare}  &  \underline{86.72±0.67} & 79.57±2.16 & 77.50±2.46    & 88.69±1.00 & 67.91±1.65 & 60.20±1.93    & 85.40±2.23 & 87.74±1.28  & 87.68±1.26  & 4.2  \\
Hype-ANE~\cite{HyperANE}    & 74.50±0.53 & 80.70±0.07 &  79.20±0.28         & 85.80±0.53 & 64.40±0.29 & 58.70±0.02        & 64.27±0.73 & \textbf{95.62±0.35} & \textbf{95.61±0.36} & 5.0 \\ 
\midrule
\textbf{\modelname~(Ours)}   & \textbf{94.00±0.63}    & \textbf{84.50±0.53} & \textbf{85.60±0.25} & \textbf{93.80±0.15} & 65.60±0.27 & 59.60±0.21 & \textbf{93.88±0.42} & 88.89±0.17 & 87.65±0.25  & \textbf{2.4} \\ 
\bottomrule
\end{tabular}}
\label{table:results}
\end{table*}

\subsection{Datasets.}
We conduct experiments on synthetic and real-world datasets to evaluate our method, and analyze model's capabilities in terms of both graph theory and real-world scenarios. 
The statistics of datasets are summarized in Table~\ref{dataset_description}.

\textbf{Synthetic Datasets.}
We generate three synthetic graph datasets using several well-accepted graph theoretical models: \textbf{Stochastic Block Model} (\textbf{SBM})~\cite{holland1983stochastic}, \textbf{Barabási-Albert} (\textbf{BA}) scale-free graph model~\cite{barabasi1999emergence}, and \textbf{Watts-Strogatz} (\textbf{WS}) small-world graph model~\cite{watts1998collective}.
For each dataset, we create 1,000 nodes and subsequently perform the graph generation algorithm on these nodes. 
For the SBM graph, we equally partition all nodes into 5 communities with the intra-class and inter-class probabilities $(p,q) = (0.21,0.025)$.
For the Barabási-Albert graph, we set the number of edges from a new node to existing nodes to a random number between 1 and 10.  
For the Watts-Strogatz graph, each node is connected to 24 nearest neighbors in the cyclic structure, and the probability of rewiring each edge is set to 0.21.
For each generated graph, we randomly remove 50\% nodes as the test set with other 50\% nodes as the positive samples and generate or sample the same number of negative samples. 

\textbf{Real-world Datasets.}
We also conducted experiments on three real-world datasets: 
\textbf{Cora}~\cite{sen2008cora} and \textbf{Citeseer}~\cite{GCN} are citation networks of academic papers;
\textbf{Polblogs}~\cite{adamic2005political} is political blogs in 2004 U.S. president election where nodes are political blogs and edges are citations between blogs.
The training settings for the real-world datasets are the same as settings for synthetic datasets. 

\subsection{Experimental Setup}
\textbf{Baselines.} 
To evaluate the proposed \modelname~, we compare it with a variety of baseline methods including:
(1) \textbf{Euclidean graph representation methods}:
We compare with other state-of-the-art unsupervised graph learning methods. GAE~\cite{kipf2016variational} and VGAE~\cite{kipf2016variational} are the autoencoders and variational autoencoder for graph representation learning; G2G~\cite{Bojchevski2018DeepGE} embeds each node of the graph as a Gaussian distribution and captures uncertainty about the node representation; 
DGI~\cite{Velickovic2019DeepGI} is an unsupervised graph contrastive learning model by maximizing mutual information. 
(2) \textbf{Euclidean graph generative adversarial networks}: GraphGAN~\cite{wang2018graphgan} learns the sampling distribution to sample negative nodes from the graph;  ANE~\cite{dai2018adversarial} trains a discriminator to push the embedding distribution to match the fixed prior; 
(3) \textbf{Hyperbolic graph representation learning}:  $\mathcal{P}$-VAE~\cite{mathieu2019poincare} is a variational autoencoder by using Poincar\'e ball model in hyperbolic geometric space; 
Hyper-ANE~\cite{HyperANE} is a hyperbolic adversarial network embedding model by extending ANE to hyperbolic geometric space. 

\begin{figure*}[!t]
\centering
\subfigure[Generalization analysis on synthetic BA and WS.]{
\includegraphics[width=1\columnwidth]{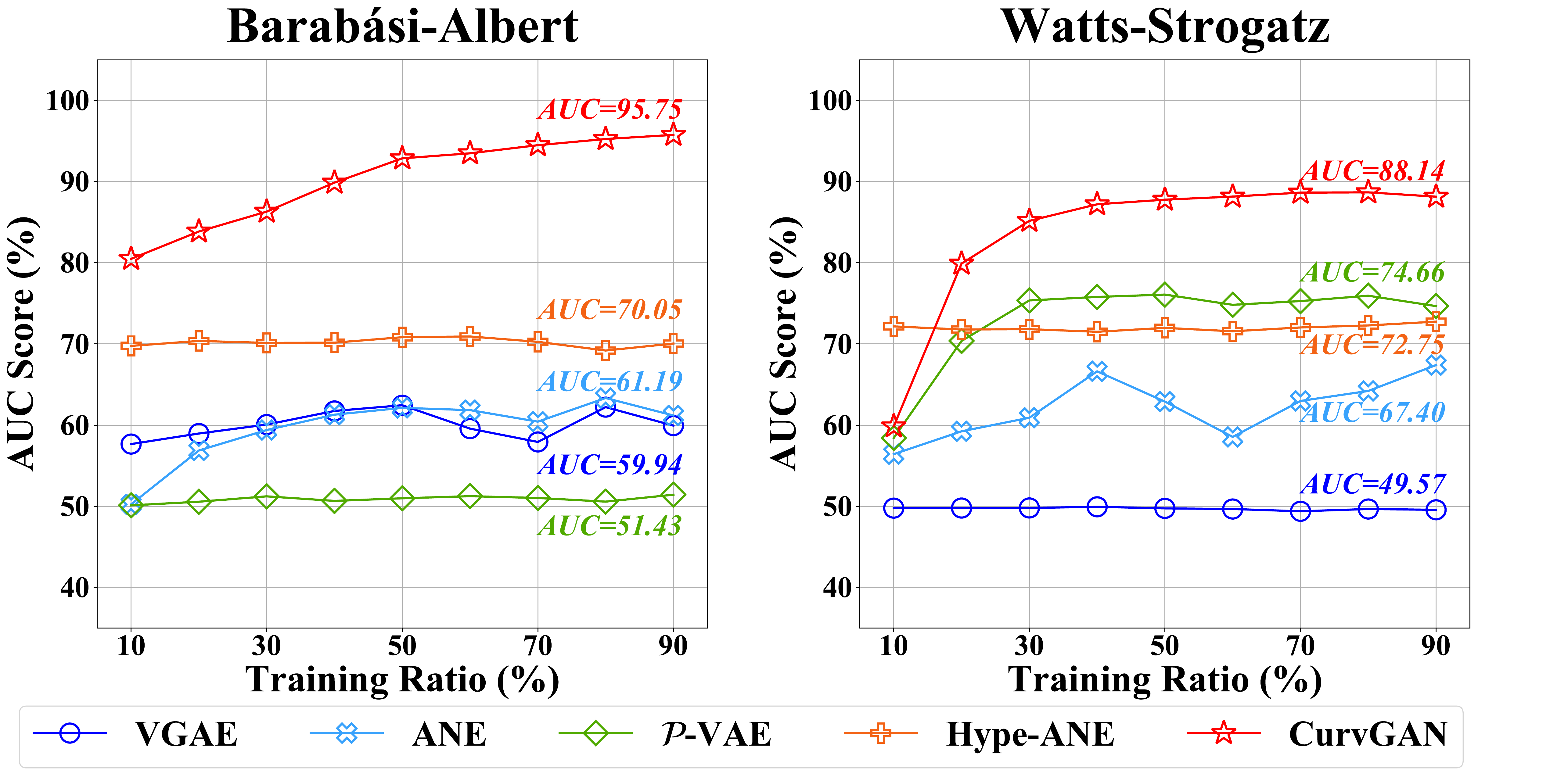} 
}
\subfigure[Robustness analysis of the edges attack on SBM.]{
\includegraphics[width=1\columnwidth]{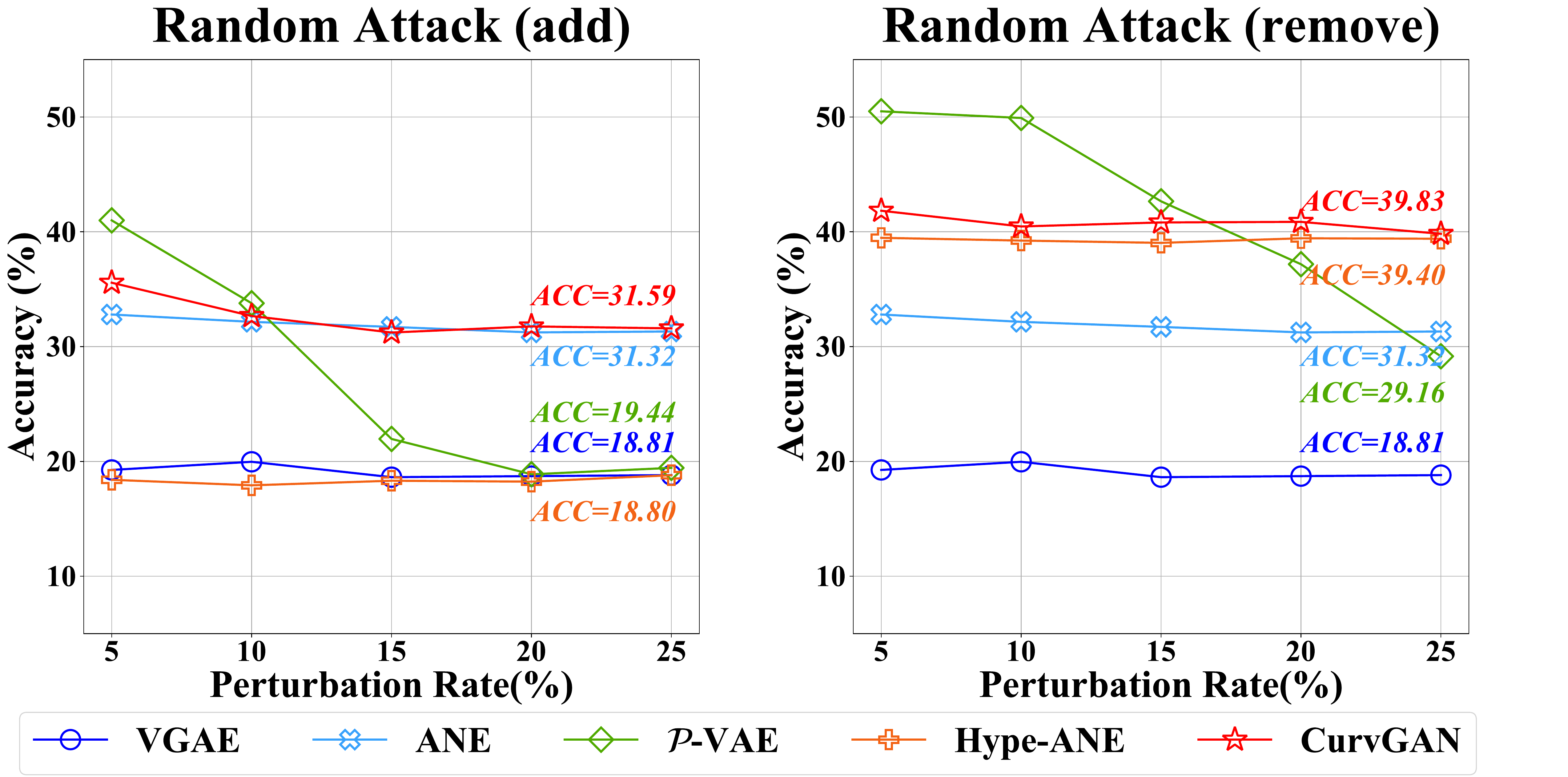} 
}
\vspace{-1em}
\caption{Generalization and robustness analysis on synthetic data.}
\vspace{-1em}
\label{fig:analysis_synth}
\end{figure*}

\begin{figure*}[ht]
\centering
\subfigure[Ground truth.]{
\includegraphics[width=0.23\linewidth]{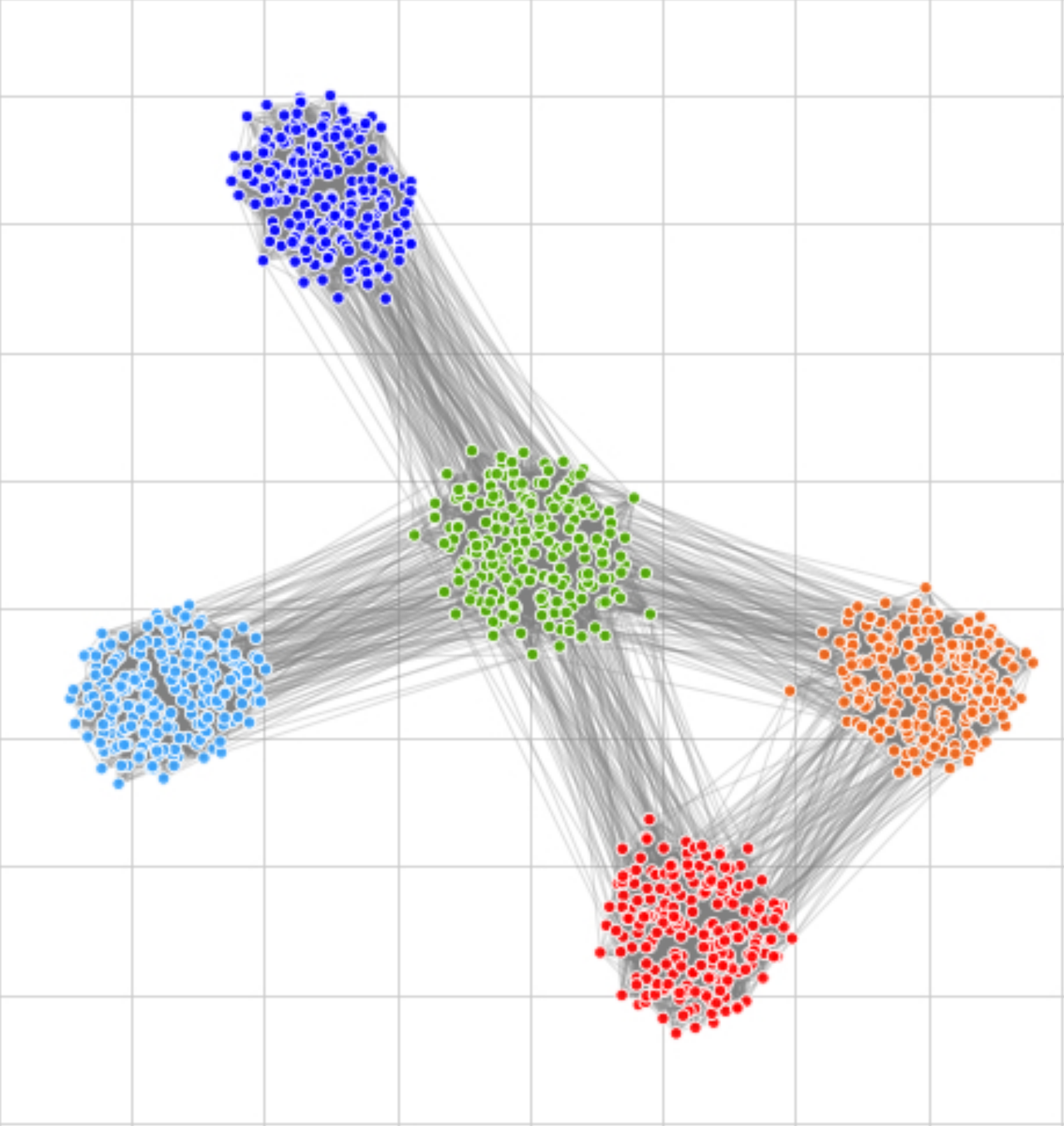}
}
\subfigure[VGAE.]{
\includegraphics[width=0.23\linewidth]{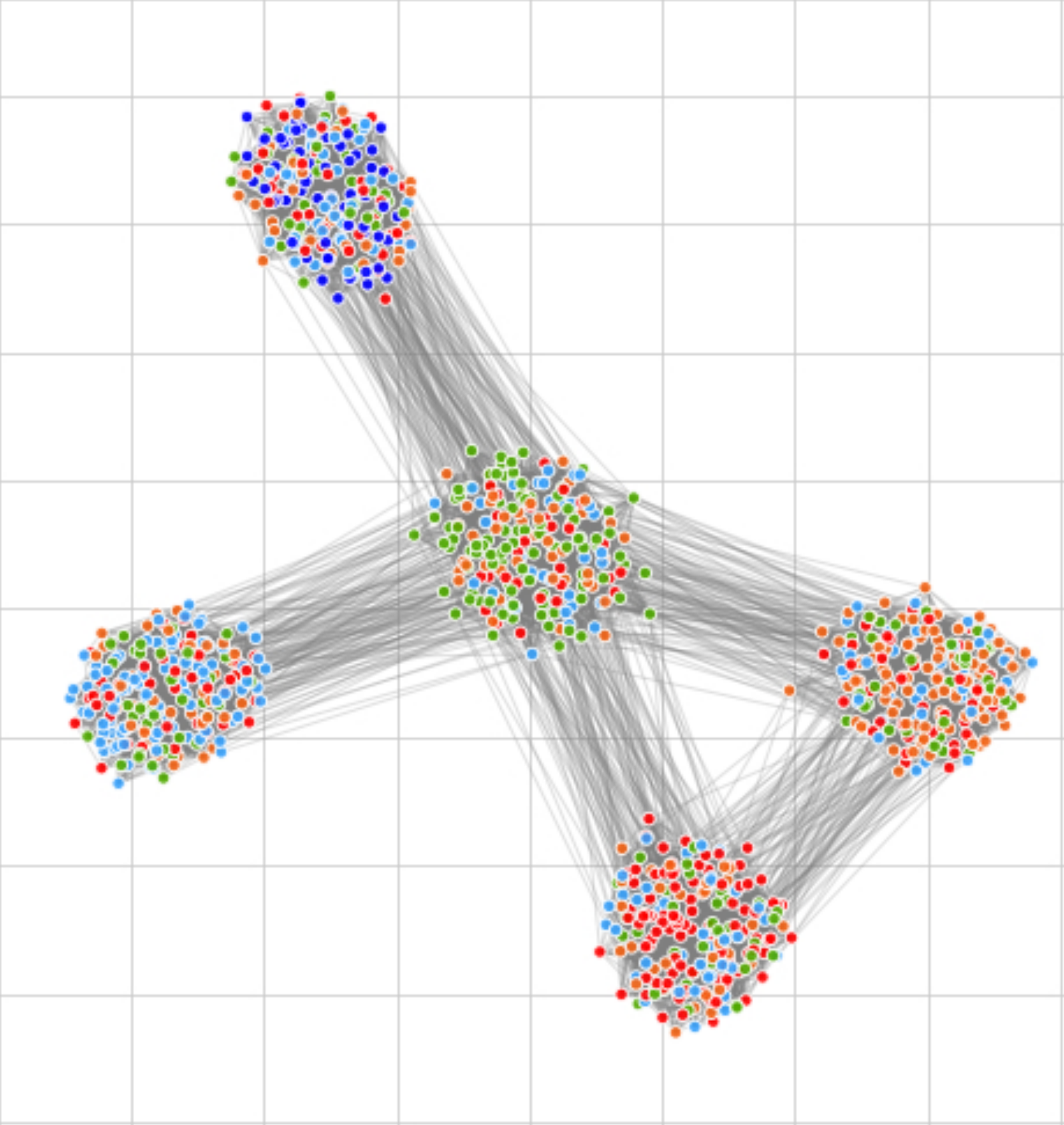}
}
\subfigure[$\mathcal{P}$-VAE.]{
\includegraphics[width=0.23\linewidth]{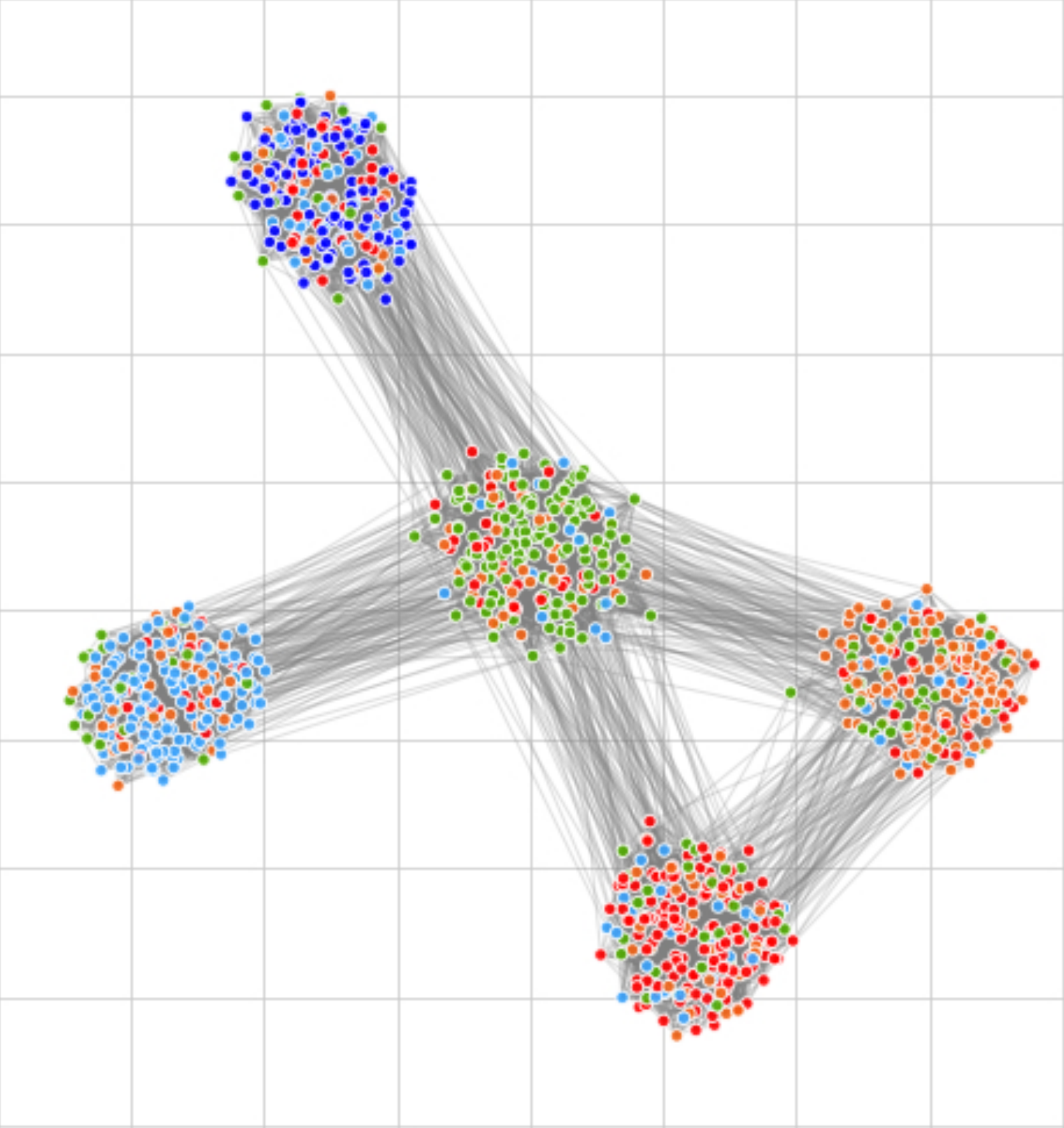}
}
\subfigure[CurvGAN.]{
\includegraphics[width=0.23\linewidth]{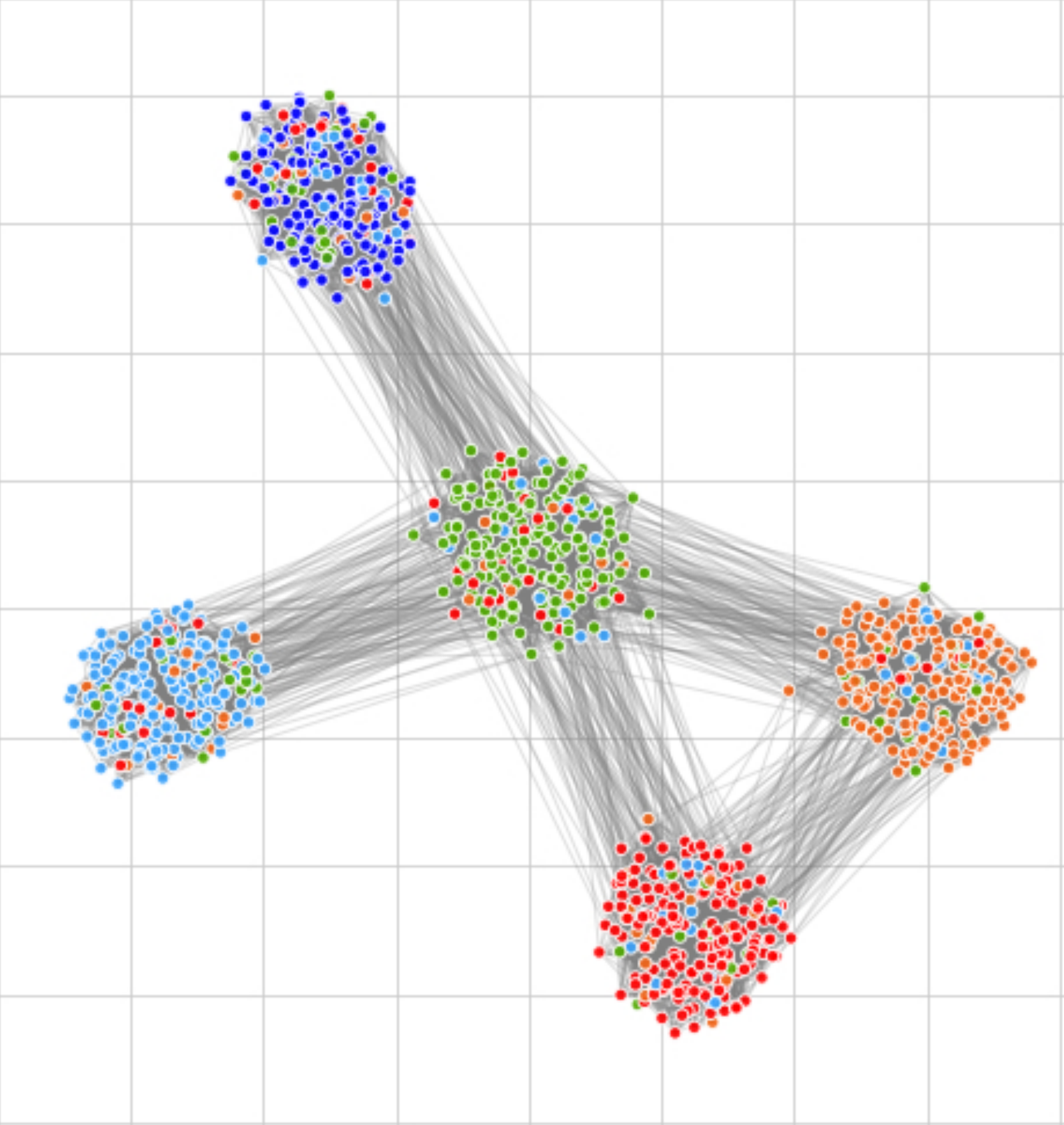}
}
\centering
\vspace{-1em}
\caption{Visualization of classification results for random attack of 10\% extra edges on SBM graph.}
\label{fig:SBM}
\vspace{-1em}
\end{figure*}

\textbf{Settings.} 
The parameters of baselines are set to the default values in the original papers. 
For \modelname, we choose the numbers of generator and discriminator training iterations per epoch $n^G_e=10, n^D_e=10$. 
The node embedding dimension of all methods is set to 16. 
The reported results are the average scores and standard deviations over 5 runs. 
All models were trained and tested on a single Nvidia V100 32GB GPU. 

\subsection{Synthetic Graph Datasets}\label{sec:synthetic}
To verify the topology-preserving capability, we evaluate our method on synthetic graphs generated by the well-accepted graph theoretical models: \textbf{SBM}, \textbf{BA}, and \textbf{WS} graphs. 
These three synthetic graphs can represent three typical topological properties: the SBM graph has more community-structure, the BA scale-free graph has more tree-like structure, and the WS small-world graph has more cyclic structure. 
We evaluate the performance, generalization and robustness of our method comprehensively on these graphs with different topological properties.

\textbf{Performance on Benchmarks.}
Table~\ref{table:synthetic} summarizes the performance of \modelname~and all baselines on the synthetic datasets.
For \textit{the link prediction task}, the performance of a model indicates the capture capability of topological properties. 
It can be observed that the hyperbolic geometric model performs better in SBM and BA graphs than in WS graphs.
The reason is that SBM and BA graphs are more "tree-like" than the WS graph. 
Our \modelname~outperforms all baselines in three synthetic graphs. 
The result shows that a good geometry prior is very important for the topology-preserving. 
CurvGAN can adaptively select hyperbolic or spherical geometric space by estimating the optimal geometric prior. 
For \textit{the node classification task}, the GAN-based methods generally outperform other methods because the synthetic graphs only have topological structure and no node features. 
The GAN-based method can generate more samples to help the model fit the data distribution. 
The results show that our CurvGAN also has the best comprehensive competitiveness in node classification benefit from the stronger negative samples. 

\textbf{Generalization Analysis.}
We evaluate the generalization capability of the model by setting different proportions of the training set. 
Figure~\ref{fig:analysis_synth} (a) shows the performances of link prediction with different training ratios. 
\modelname~ significantly outperforms all baselines, even when the training ratio is small. 
In addition, \modelname gains more stable performances than other GAN-based methods across all datasets, which demonstrates the excellent structure-preserving capability of the network latent geometric space. 
We observe an interesting phenomenon: the non-Euclidean geometry models have very smooth performances on the synthetic graphs with a single topological property.
It demonstrates again that a single negative curvature geometry lacks generalization capability for different graph datasets. 

\begin{figure*}[!t]
\centering
\subfigure[Generalization analysis on \textbf{Cora} and \textbf{Citeseer}.]{
\includegraphics[width=1\columnwidth]{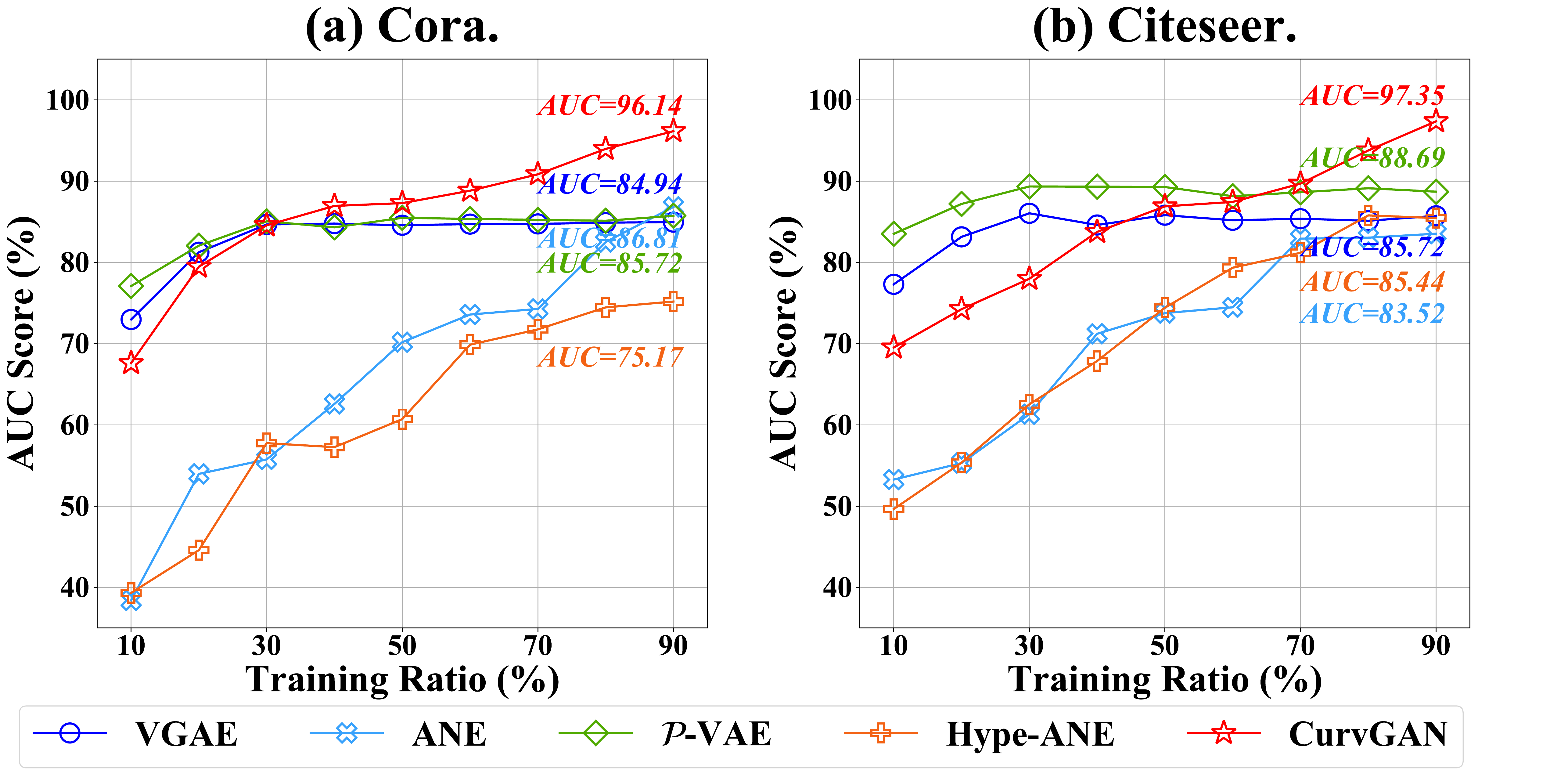} 
}
\subfigure[Robustness analysis of the edges attack on \textbf{Cora}.]{
\includegraphics[width=1\columnwidth]{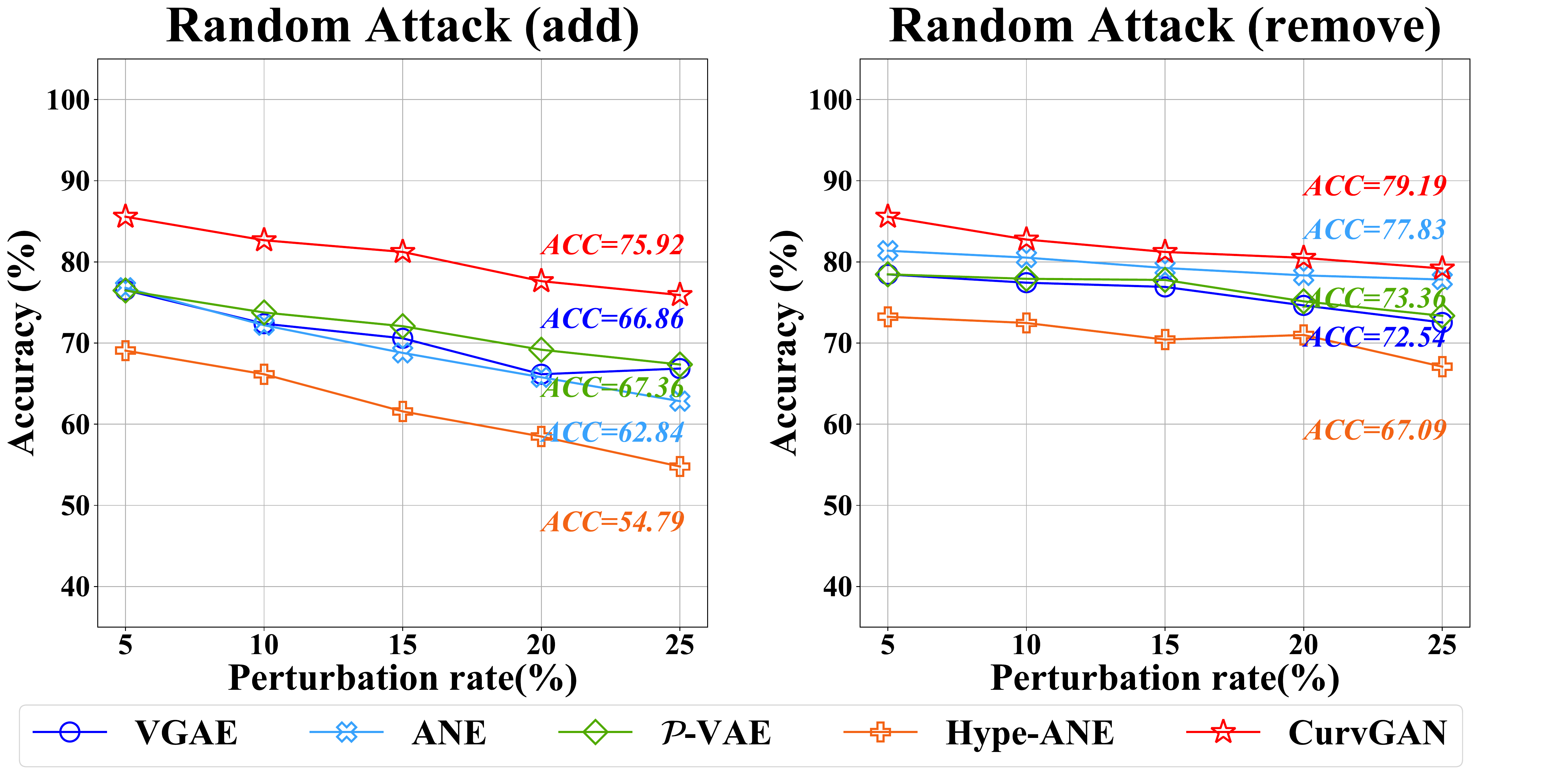} 
}
\vspace{-1em}
\caption{Generalization and robustness analysis on real-world data.}
\label{fig:analysis_real}
\end{figure*}

\begin{figure}[!t]
\centering
\includegraphics[width=1\linewidth]{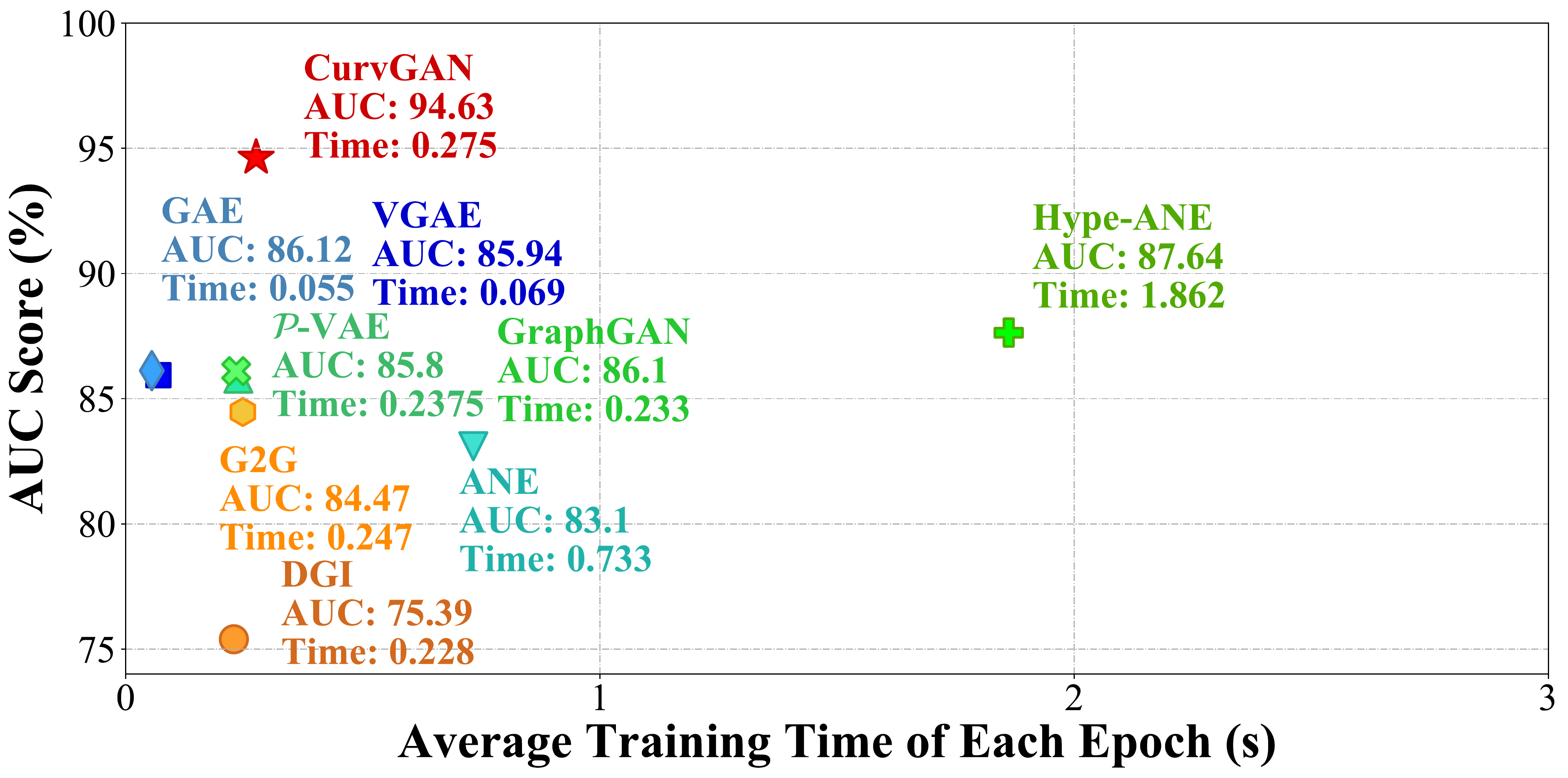}
\vspace{-1em}
\centering
\caption{Training efficiency analysis. }
\label{fig:efficiency}
\end{figure}

\textbf{Robustness Analysis.}
Considering a poisoning attack scenario, we leverage the RAND attack, provided by DeepRobust~\cite{li2020deeprobust} library, to randomly add or remove fake edges into the graph.
Specifically, we randomly remove and add edges of different ratios (from 5\% to 25\%) as the training data respectively, and randomly sample 10\% nodes edges from the original network as the test data. 
The results are shown in Figure~\ref{fig:analysis_synth} (b). 
\modelname~consistently outperforms Euclidean and hyperbolic models in different perturbation rates. 
Figure~\ref{fig:SBM} shows the visualization of an edge attack scenario. 
It can observe that $\mathcal{P}$-VAE has no significant improvement than Euclidean VGAE. 
Since noisy edges may perturb some tree-like structures of the original SBM graph, leading to hyperbolic models no longer suitable for the perturbed graph topology. 
Overall, our CurvGAN has significant advantages in terms of robustness. 

\subsection{Real-world Graph Datasets}
To verify the practicality of our model, we evaluate \modelname~in terms of performance, generalization, and robustness on real-world datasets for two downstream tasks: link prediction and node classification. 
As in Section~\ref{sec:synthetic}, we use the same unsupervised training setup on three real-world datasets, \textbf{Cora}, \textbf{Citseer}, and \textbf{Polblogs}. 
In addition, we also analyze the computational efficiency of \modelname~and all baselines. 

\textbf{Performance on Benchmarks.}
Table~\ref{table:results} summarizes the performance of \modelname~ and all baselines on three real-world datasets.
For \textit{the link prediction task}, our CurvGAN outperforms all baselines (including hyperbolic geometry models) in real data and can learn better structural properties based on correct topological geometry priors.
In contrast to a single hyperbolic geometric space, a unified latent geometric space can improve benefits for learning better graph representation in real-world datasets with complex topologies.
For \textit{the node classification task}, we combine the above link prediction objective as the regularization term in node classification tasks,  to encourage embeddings preserving the network structure. 
Table~\ref{table:results} also summarizes Micro-F1 and Macro-F1 scores of all models on three real-world datasets.
It can be observed that the Euclidean models have comparative performance. 
The reason is that the node labels are more dependent on other features (e.g. node's attributes or other information) than topological features. 

\textbf{Generalization Analysis.}
Figure~\ref{fig:analysis_real} (a) shows the performances of link prediction with different training ratios. 
\modelname~ significantly outperforms all baselines even when the training ratio is small. 
In addition, we find that the stability of the autoencoders VGAE and $\mathcal{P}$-VAE is higher than the GAN-based methods (ANE, GraphGAN, Hype-ANE and our CurvGAN), although their performances are outperformed by CurvGAN rapidly.
The reason is the GAN-based method needs more samples to fit the data distribution.

\textbf{Robustness Analysis.}
To evaluate the robustness of the model, we also perform a poisoning attack RAND by DeepRobust~\cite{li2020deeprobust} on the real-world data, and the setting is the same as in the robustness analysis in Section~\ref{sec:synthetic}.
Figure~\ref{fig:analysis_real} (b) shows that \modelname~and all baselines under the edges attack scenario on Cora. 
Our \modelname~always has better performance even when the network is very noisy. 
Riemannian geometry implies the prior information of the network, making \modelname~has the excellent denoising capability.

\textbf{Efficiency Analysis.}
Figure~\ref{fig:efficiency} illustrates the training time of \modelname~ and baselines on Cora for link prediction. 
\modelname~ has both the best performance and second-best efficiency. 
It can be observed that our CurGAN has the best computational efficiency in the GAN-based method (ANE, GraphGAN, and Hype-ANE).
In general, the results show that the comprehensive evaluation of our model outperforms baselines on all datasets, which indicates that the network latent geometry can significantly improve the computational efficiency and scalability of network embedding.

\section{Conclusion}
In this paper, we proposed \modelname, a novel curvature graph generative adversarial network in Riemannian geometric space.
\modelname~is the first attempt to combine the generative adversarial network and Riemannian geometry, to solve the problems of network discrete topology representation and topological heterogeneity. 
For graphs with different topological properties, \modelname~can effectively estimate a constant curvature to select an appropriate Riemannian geometry space, and leverage Ricci curvature regularization for capturing local structure. 
Moreover, \modelname's generator and discriminator can perform operations and continuous computation for any node-pairs in the latent geometric space, leading to good efficiency and scalability. 
The experimental results on synthetic and real-world datasets demonstrate that \modelname~consistently and significantly outperforms various state-of-the-arts unsupervised methods in terms of performance, generalization, and robustness.

\begin{acks}
The authors of this paper were supported by the NSFC through grants (No.U20B2053), and the ARC DECRA Project (No. DE200100964). The corresponding author is Jianxin Li.
\end{acks}

\bibliographystyle{ACM-Reference-Format}
\bibliography{reference}

\end{document}